\documentclass[11pt]{article}

\usepackage[toc]{appendix}
\usepackage{minitoc}
\usepackage{hyperref}       
\hypersetup{
	colorlinks=true,       
	linkcolor=blue,        
	citecolor=blue,        
	filecolor=magenta,     
	urlcolor=blue
}

\noptcrule

\usepackage[utf8]{inputenc}
\usepackage{authblk}
\usepackage{setspace}
\usepackage{url}
\usepackage{graphicx}
\graphicspath{ {../} }
\usepackage{subcaption}
\usepackage{amsmath}
\usepackage{booktabs}
\usepackage{wrapfig}
\usepackage{enumitem}

\graphicspath{ {../figures/} }

\usepackage{amssymb, amsmath, amsthm,mathtools}
\newtheorem{theorem}{Theorem}[section]

\newtheorem{lemma}[theorem]{Lemma}

\theoremstyle{remark}
\newtheorem{remark}{Remark}[section]

\newtheorem{conjecture*}{Conjecture}
\theoremstyle{plain}
\DeclareMathOperator*{\argmax}{arg\,max}

\newcommand{\E}{\mathbb{E}}
\newcommand{\R}{\mathbb{R}}
\newcommand{\bbP}{\mathbb{P}}

\newcommand{\calN}{\mathcal{N}}

\newcommand{\calX}{\mathcal{X}}

\newcommand{\calG}{\mathcal{G}}

\newcommand{\ARL}{\mathrm{ARL}}

\newcommand{\ONNC}{\mathrm{ONNC}}
\newcommand{\ONNR}{\mathrm{ONNR}}

\usepackage{color,xcolor}

\title{Neural network-based CUSUM for online change-point detection}
\author[1]{Tingnan Gong}
\author[1]{Junghwan Lee}
\author[2]{Xiuyuan Cheng}
\author[1]{Yao Xie}

\affil[1]{
{\small H. Milton Stewart School of Industrial and Systems Engineering,
Georgia Institute of Technology}}
\affil[2]{
{\small 
Department of Mathematics, Duke University}}

\date{}

\usepackage[margin=1in]{geometry}

\begin{document}

\maketitle


\begin{abstract}
Change-point detection, detecting an abrupt change in the data distribution from sequential data, is a fundamental problem in statistics and machine learning. CUSUM is a popular statistical method for online change-point detection due to its efficiency from recursive computation and constant memory requirement, and it enjoys statistical optimality. CUSUM requires knowing the precise pre- and post-change distribution. However, post-change distribution is usually unknown a priori since it represents anomaly and novelty. Classic CUSUM can perform poorly when there is a model mismatch with actual data. While likelihood ratio-based methods encounter challenges facing high dimensional data, neural networks have become an emerging tool for change-point detection with computational efficiency and scalability. In this paper, we introduce a neural network CUSUM (NN-CUSUM) for online change-point detection. We also present a general theoretical condition when the trained neural networks can perform change-point detection and what losses can achieve our goal. We further extend our analysis by combining it with the Neural Tangent Kernel theory to establish learning guarantees for the standard performance metrics, including the average run length (ARL) and expected detection delay (EDD). The strong performance of NN-CUSUM is demonstrated in detecting change-point in high-dimensional data using both synthetic and real-world data.\footnote{First two authors are equal contribution. 
Corresponding author Yao Xie (yao.xie@isye.gatech.edu).}
\end{abstract}



\setcounter{page}{1}

\section{Introduction}
Quick detection of unknown changes in the distribution of sequential data, known as change-point detection, is a crucial issue in statistics and machine learning \cite{siegmund1985sequential,tartakovsky2014sequential,poor2008quickest,xie2021sequential}. It is often triggered by an event that may lead to significant consequences, making timely detection critical in minimizing potential losses. For instance, a change-point could arise when production lines become unmanageable. Early detection and remediation by practitioners would result in lesser loss. With the rapidly growing amount of streaming data, the change-point detection applies to practical problems across diverse fields such as gene mapping \cite{siegmund2007statistics}, sensor networks \cite{poor2008quickest}, brain imaging \cite{barnett2016change}, and social network analysis \cite{peel2015detecting,li2017detecting}.


In the modern high-volume and complex data era, across numerous applications, sequential data is collected over networks with higher dimensionality and more complex distributions than before. Therefore, the conventional setting of the change-point detection problem has been surpassed in scope by its modern version. The assumptions in the classical methodologies are possibly too restrictive to tackle the newly occurring challenges, including the high-dimensionality and complex spatial and temporal dependence of sequential data. 

Various methods have been proposed to address the challenges above. Deep learning architectures were often used to learn complex spatio-temporal dependence in the data for density estimation (e.g., \cite{liu2021density}) for change-point detection~\cite{
gupta2022real}. Moustakides and Basioti~\cite{moustakides2019training}
aimed to develop a suitable optimization framework for training neural network-based estimates of the likelihood ratio or its transformations where the estimates can be used for change-point detection and hypothesis testing. Detecting changes by statistics generated by online training of neural networks has been recently studied in a few previous works. Hushchyn et al.~\cite{hushchyn2020online} introduce online neural network classification (ONNC) and online neural network regression (ONNR), which is based on sequentially applying neural networks for classification or regression for comparing the current batch of data with the immediate past batch of data, to obtain the detection statistic with separating training and testing. However, prior works did not present theoretical guarantees of the algorithms; here, we generalize the prior works by considering general losses, presenting a formal theoretical guarantee with extensive numerical experiments. Moreover, ONNC and ONNR \cite{hushchyn2020online} can be viewed as Shewhart chart type (without accumulating history information and with short memory) rather than a CUSUM-type of algorithm (accumulating all past history information through the recursion); it is known that Shewhart charts may be good at detecting larger changes but not small changes, while CUSUM procedures are known to be sensitive to small changes and enjoy asymptotic optimality properties. 

Recently, \cite{li2022automatic} presented a novel {\it offline} detection method based on training a neural network and a theory that quantifies the error rate for such an approach. The paper demonstrates competitive empirical performance with the standard CUSUM-based classifier for detecting a change in mean. In contrast, our NN-CUSUM aims to detect an abrupt change quickly online from sequential data, using the CUSUM recursion procedure, which differs from \cite{li2022automatic}; our numerical experiments also studied detecting general types of changes.  

In this paper, we leverage neural networks to efficiently detect the change-point in the streaming data with high-dimensionality and complex spatial-temporal correlations. The starting point is to convert online change-point detection into a classification problem, of which neural networks are highly capable. Now that the explicit form of the distribution and the log-likelihood ratio are unknown, neural networks benefit us in that the training loss of the neural networks will approximate the underlying true log-likelihood ratio between pre- and post-change distributions. Further, we couple the loss from the neural networks with CUSUM recursion. We present the statistical performance analysis for the average run length (ARL) and the expected detection delay (EDD) for the proposed algorithm, combining stopping time analysis and neural network training analysis through neural-tangent-kernel (NTK). We summarize the idea as an efficient online training procedure and design extensive simulations and empirical experiments to thoroughly discuss the proposed method against both conventional and state-of-the-art methods. Initial results were presented in our earlier paper \cite{lee2023training}; the current version includes more in-depth theoretical analysis and comprehensive numerical study and comparison. 

\section{Preliminaries}\label{sec:format}
Consider a sequence of observations $\{x_t\in\mathbb R^d:t=1,2,\ldots\}$. At some point, there may be a change-point $k$ such that the distribution of the data shifted from $f_0$ to $f_1$, which can be formulated as the following composite sequential hypothesis test
\begin{equation}
    \begin{split}
        H_0: & \quad x_1, x_2,\ldots, x_t, \ldots  \overset{\mathrm{iid}}{\sim} f_0, \\
        H_1: & \quad x_1, x_2, \ldots, x_k \overset{\mathrm{iid}}{\sim} f_0, \quad x_{k+1}, \ldots x_t, \ldots \overset{\mathrm{iid}}{\sim} f_1, 
    \end{split}
\end{equation}
where $f_0$ and $f_1$ represent the pre- and post-change distributions, respectively. The expectations under $f_0$ or $f_1$ are denoted by $\E_0[\cdot]$ and $\E_1[\cdot]$, respectively. We assume that pre-change data under $f_0$ are adequate to converge any interested pre-change statistics. We call the pool of pre-change data in the training process as {\it reference sequence}. This assumption commonly holds since, in practice, the process is mostly pre-change, which allows the sufficient collection of reference data to represent the ``normal'' state. 
Another assumption, which can be considered equivalent to sufficient pre-change data up to some estimation error, is that the pre-change distribution $f_0$ is known.

The classic CUSUM procedure is popular, possibly due to its computationally efficient recursive procedure, and enjoys various optimality properties (e.g., a survey \cite{xie2021sequential}). CUSUM detects the change using the exact likelihood ratio through the recursion
$
S_t^{\rm C} = \left(S_{t-1}^{\rm C} + \log [f_1(x_t)/f_0(x_t)] \right)^+.
$
where $S_0^{\rm C} = 0$, and $(x)^+ = \max\{x, 0\}$. CUSUM procedure raises an alarm at a stopping time 
$
\tau_{\rm C} = \inf\{t: S_t^{\rm C} > b\},
$
where $b>0$ is the pre-specified threshold. However, the post-change distribution $f_1$ is usually unknown and in practice. CUSUM may suffer performance degradation when the assumed distributions deviate from the true distributions. 

\section{Neural network CUSUM}\label{sec:NNCUSUM} 

This section proposes the Neural Network (NN)-CUSUM procedure. We consider a neural network function $g_\theta( x): \R^d\rightarrow \R$, parameterized by $\theta$. The network sequentially takes a pre-change distributed reference sequence and an unknown distributed online sequence as input. 

During training, we convert the change-point detection problem into a binary classification problem. At time index $i$, we label data from the reference and online sequences with $y_i=0$ and $y_i=1$, respectively. We feed the labeled data into the neural network and train it using a carefully crafted logistic loss function (such as \cite{cheng2022classification}). Subsequently, the neural network computes the logistic loss for each data point in the reference and online sequences for every time step. We employ the difference between the average logistic function values of the reference and online sequences as the sequential statistics for detecting change-points. Figure~\ref{fig:data_process} visualizes the proposed framework. 

The stopping time $\tau$ is crucial in the sequential hypothesis testing approach for detecting changes in a streaming dataset. It indicates the time to terminate the data stream and declares the occurrence of a distribution shift before time $\tau$. We adopt the CUSUM recursion in our proposed scheme to construct the stopping time but replace the exact log-likelihood ratio with an approximation from a neural network. Before presenting the specifics of our approach, we provide an overview of the exact CUSUM procedure and other relevant methods.

\subsection{Training}
\begin{figure}[t]
\begin{center}
\includegraphics[width = 0.48\textwidth]{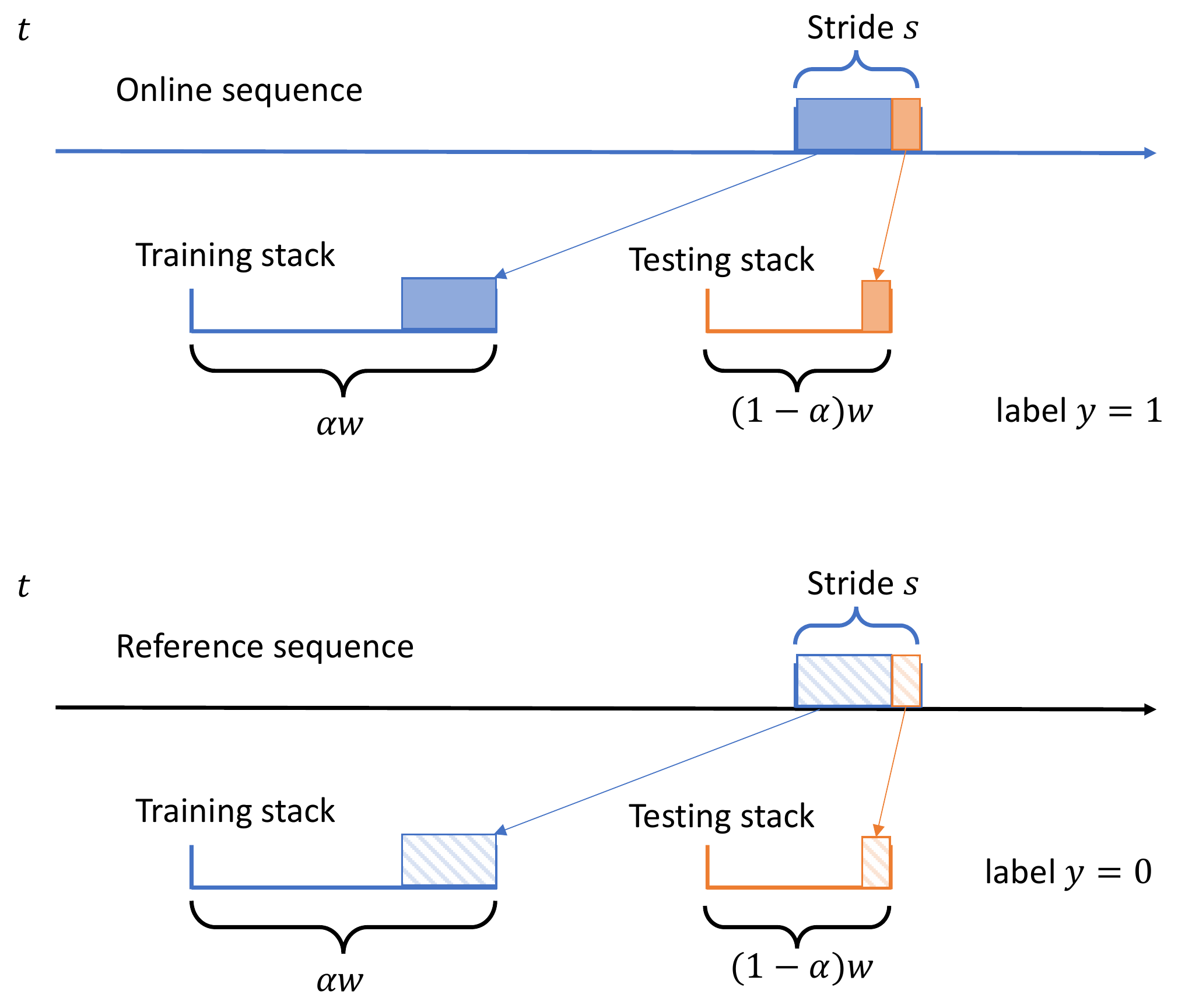} \quad
\includegraphics[width = 0.48\textwidth]{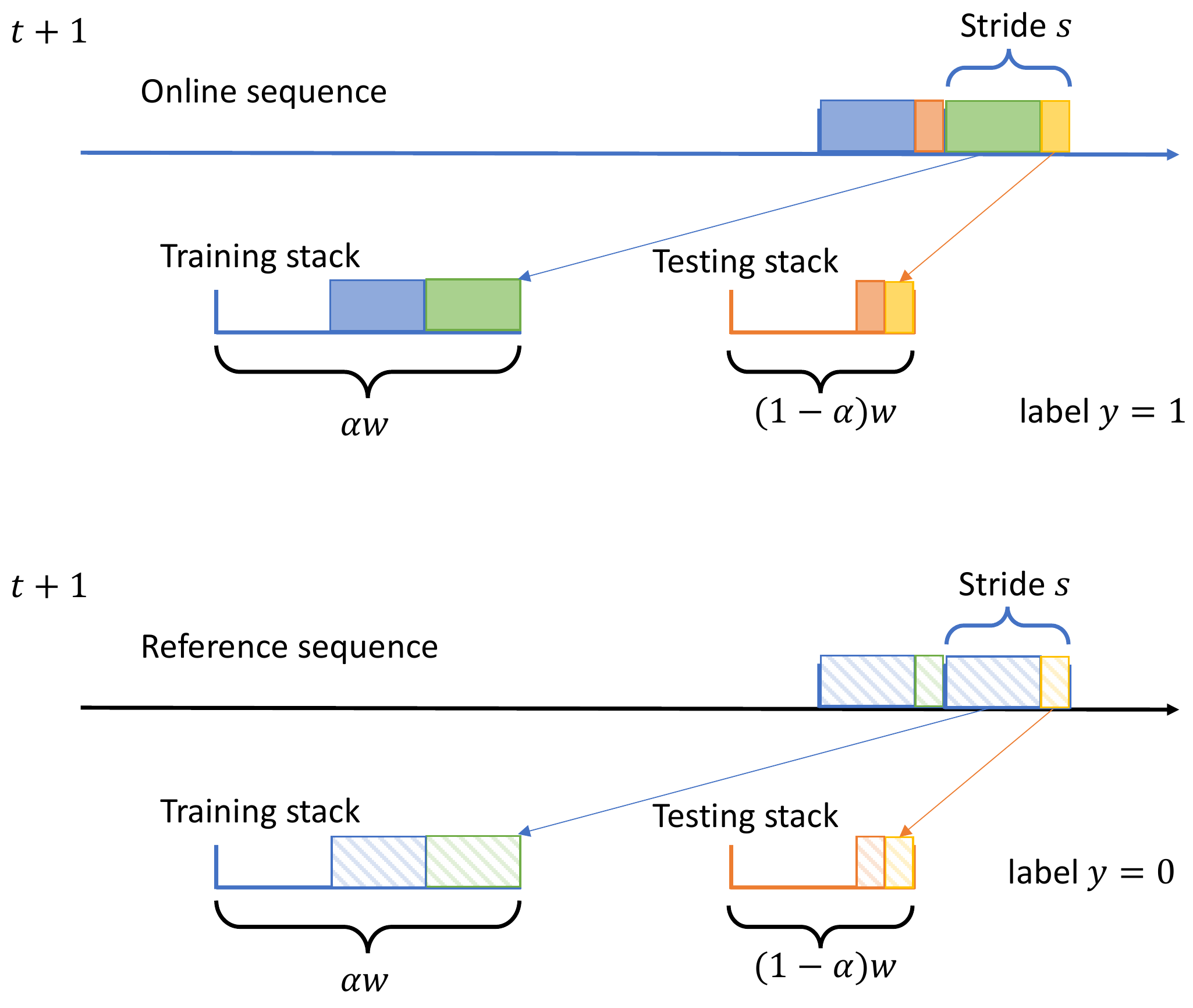}
\end{center}
\vspace{-0.125in}
\caption{
Illustration of the NN-CUSUM procedure, running from $t$ (left) to $t+1$ (right).
When the newest batch from the streaming data (treated with label $y_i = 1$) is received, it is divided to put into the training stack and the testing stack, respectively (the oldest samples are purged from the stacks to keep a constant stack size). 
The neural network inherited from the previous $t$ will be updated using a stochastic gradient descent algorithm for {\it one-pass} through the freshened training stack data. The updated neural network is then used to compute the test statistic for data in the freshened {\it testing stack} data. The label $y_i=0$ (from reference samples) is constructed completely in parallel. 
The training/testing stack sizes are hyper-parameters to be tuned in practice to achieve a good {\it detection performance}.
}
\label{fig:data_process}
\end{figure}

The neural network training is conducted via SGD updates of the neural network parameter $\theta$ on the training stack. As is shown in Figure~\ref{fig:data_process}, the training stack contains multiple mini-batches of data in $[t-w, t]$ window from the arriving data stream. Within the window, we use a fraction of $\alpha$ samples for training and a fraction of $1-\alpha$ samples for testing. This gives the online training data set $\widetilde{X}_t^{\rm tr}$ and the online test data set $\widetilde{X}_t^{\rm te}$, 
with a size of $\alpha w$ and $(1- \alpha)w$, $\alpha \in (0, 1)$, respectively. The window length $w$ and training split fraction $\alpha$ are algorithm parameters.

The sliding window moves forward with a stride $s$.  
Half of the stride 
is put into the training stack, and the other half is put into the testing stack, and the data in the online sequence 
are labeled 1, and the data in the reference sequence 
are labeled 0. We update the reference sequence by drawing randomly from reference samples.  

Let $\varphi (x, \theta)$ (to be related to $g_\theta(x)$ later) 
be the neural network function that maps from $\calX \subset \R^d \to \R$ 
and $\theta$ is the network parameter. Consider training the neural network using the following  general loss, which is computed from the combined training split of arrival and reference samples on the window
\[D_t^{\rm tr}:= X_t^{\rm tr} \cup \widetilde X_t^{\rm tr} = \{ x_i \}_{i=1}^{2 m},\] $m =  \alpha w$, 
labeled by $y_i =  1 $ if $x_i \in X_t^{\rm tr}$ and $y_i = 0$ if $x_i \in \widetilde X_t^{\rm tr} $,
\begin{equation}\label{eq:loss-theory-general}
\ell( \theta; D_t^{\rm tr}) 
= \frac{1}{m}\sum_{i=1}^{2m} l( \varphi (x_i, \theta) , y_i),
\end{equation}
The per-sample loss function $l(u,y)$ can be chosen to be different losses:
\begin{itemize}
\item For the logistic loss, we have
\begin{equation}\label{eq:def-l-logit}
l_{\rm logit}( u, y )= y \log( 1+e^{-u}) + (1-y) \log(1+e^u).
\end{equation}
The choice of the loss function is motivated by the fact that the functional global minimizer gives the log density ratio: 
For $f_0$ and $f_1$, which are probability densities on $\R^d$, let
\[
\ell [ \varphi ] = \int \log (1+e^{ \varphi(x)}) f_0(x) dx + \int \log (1+e^{- \varphi (x)}) f_1(x) dx,
\]
which is the population loss, 
then $\ell[  \varphi]$ is minimized at $ \varphi^* = \log(f_1/f_0)$. 
One can verify that the functional $\ell [ \varphi]$  is convex with respect to the perturbation in $ \varphi$. By that 
\[\frac{\delta \ell }{ \delta  \varphi}
= \frac{f_0 e^ \varphi -f_1}{1+e^\varphi},\]
we know $\delta \ell / \delta  \varphi$ vanishes when $e^ \varphi = f_1/f_0$, that is $ \varphi=  \varphi^*$, and this is a global minimum of $\ell[ \varphi]$. Similar observation has also made been by   \cite{moustakides2019training}, which shows by training neural network according to a certain loss lead leads to log-likelihood function in theory.
\item Apart from the logistic loss, we also consider the $\ell^2$-loss, defined as 
\begin{equation}\label{eq:def-l-l2}
l_{2}( u, y )= \frac{1}{2}\left[ y (u-1)^2
    + (1-y) (u+1)^2\right].
\end{equation}
\item The linearized version of the logit loss, motivated by the MMD test statistic \cite{cheng2021neural}, as
\begin{equation}\label{eq:def-l-mmd}
l_{\rm MMD}( u, y )= -y u + (1-y) u.
\end{equation}
\end{itemize}
Here, we assume that the SGD training to minimize \eqref{eq:loss-theory-general} is conducted on $D_t^{\rm tr}$ for each window indexed by $t$ separately from scratch. At time $t$, the trained network parameter is denoted as $\hat{\theta}_t$, and the trained neural network is $\varphi(\cdot, \hat \theta_t)$.

In practical implementation, the neural network parameter $\theta$ is initialized at time zero and then updated by stochastic gradient descent (SGD) (e.g., Adam \cite{kingma2014adam}) through batches of training samples in a sliding training window. 
Newly arrived data stay in the stack for $w/s$ steps.
We trained the network on the training stack in every $k$ step. Thus, each data sample has been used $(w/s)/k$ times for training $\theta$. The number $(w/s)/k$ can be viewed as a hyperparameter indicating an effective number of epochs.

\subsection{Testing}

At each time $t$, the trained neural network provides a test function $g_{\hat{\theta}_t}(x)$. 
\begin{itemize}
\item For logistic and $\ell_2$ loss, we set the function 
$g_{\hat{\theta}_t} = \varphi (\cdot, \hat{\theta}_t)$.
\item For MMD loss,  $g_{\hat{\theta}_t}$ will be specified in Section \ref{sec:NTK}:
\begin{equation}
    g_{\hat{\theta}_t }(x) = \frac{1}{\gamma} (  \varphi (x, {\hat{\theta}_t}) - \varphi  ( x, {{\theta_0}} ) ).
    \label{g_def_MMD}
\end{equation}
where the $\gamma > 0$ is the learning rate \cite{cheng2021neural}.
\end{itemize}
%

To perform online detection, we compute the sample average on the testing windows  $X^{\rm te}_{t}$
and $\widetilde X^{\rm te}_{t}$
(recall that $\tilde X_t \sim f_0$ are drawn sequentially from the reference samples):
\begin{equation}
\eta_t 
: = \frac{1}{(1-\alpha) w}
\left[\sum_{x\in X^{\rm te}_t} g_{\hat{\theta}_t}(x)
    - \sum_{ x \in \widetilde X^{\rm te}_{t}} g_{\hat{\theta}_t}(x)\right].
    \label{eta_def}
\end{equation}
Recall that $(1-\alpha) \in (0, 1)$ describes the proportion of the data in the window used for testing; so $\alpha w$ data are used for updating the training stack. 

\subsection{NN-CUSUM}

Starting with $S_0 = 0$, the recursive CUSUM is computed as 
\begin{equation}\label{eq:NNCUSUM_recursion}
    S_t = \max\{S_{t-1} + \eta_t - D, 0\}, t= 1, 2 \ldots
\end{equation}
where $D\geq 0$ is a small constant called the ``drift'' to ensure that the expected value of the drift term is negative before the change and positive after the change; we will give the condition of the test function that can satisfy this in the following section. For some pre-specified threshold $b >0$,
the detection procedure is specified by the stopping time
\begin{equation}\label{eq:NNCUSUM}
    \tau = \inf\{t: S_t > b\}.
\end{equation}

In practical implementations, for simplicity, our algorithm computes $\eta_t$ at intervals of $s$ (for ``stride''), which is the size of mini-batches of loading the stream data, e.g. $s=10$; in the extreme case, $s =2$, since we need half-mini-batch split into stacks. When $s = w$, the sliding window contains completely non-overlapping data. The test statistic $S_t$ is computed with the stride $s$. 

\begin{figure}[t]
\begin{center}
\includegraphics[width = 0.6\textwidth]{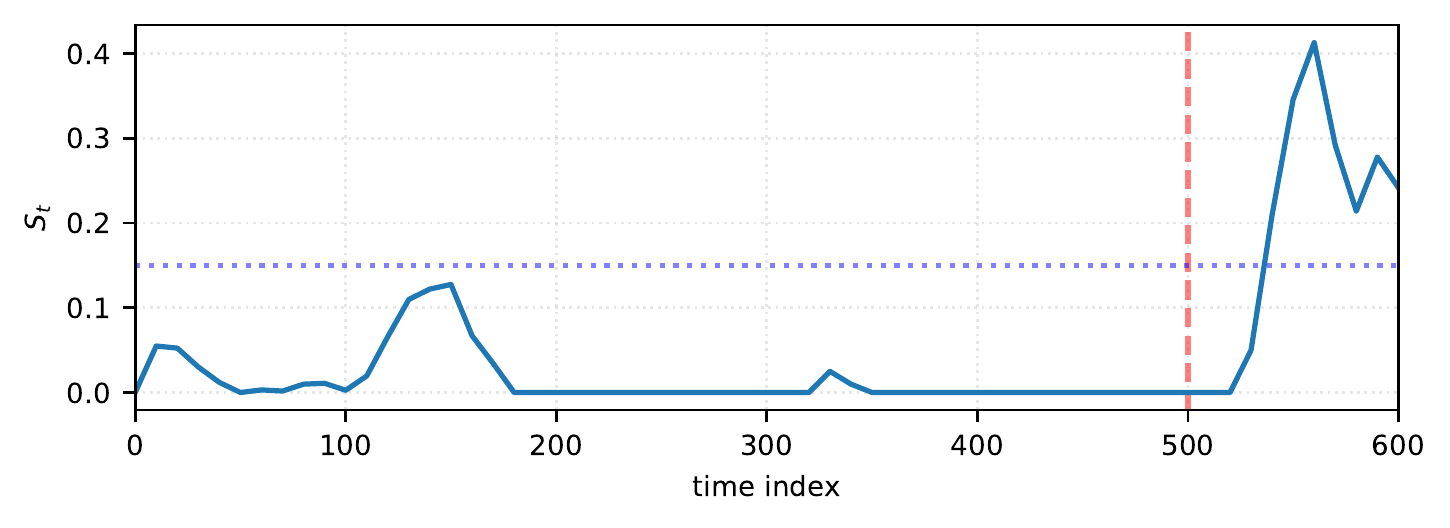}
\end{center}
\caption{One sampled trajectory of test statistics $S_t$ of NN-CUSUM on sequential detection of Higgs boson~\cite{baldi2014searching}. The red dashed line denotes the change point when signal changes from background signal to Higgs boson producing signal. The blue dotted line denotes the approximate threshold for detection; this shows that we can choose a threshold to detect the change quickly after it has occurred.}
\label{fig:sample_trajectory}
\end{figure}

\section{Theoretical analysis}

This section will explain the rationale for designing the algorithm and why it can detect the change points under some general conditions.
Below, we use $\mathbb E_0$  to denote the expectation of data when there is no change, i.e., all the data are i.i.d. with the distribution $f_0$, and $\mathbb E_1$ to denote the expectation when the change happens at the first time instance, i.e., all the samples are i.i.d. with the distribution $f_1$. 

\subsection{Why this works?}\label{sec:theory justification}

First, we show what test function $g_\theta(x)$ is necessary to achieve change-point detection. For CUSUM detection to work, we need the property that the expected value of the increment is negative before and positive after the change. Due to this property, the detection statistics, before the change-point, can stay close to 0 and rarely cross the threshold to avoid false alarms. After the change, the detection statistic can rise quickly and cross the threshold to detect the change. Admittedly, merely \eqref{g_def} is not sufficient for gaining a detection power, as the mean, variance, and even other higher-order moments matter. Below, we analyze the property of the expected increment to shed light on the above.

The increment in CUSUM \eqref{eq:NNCUSUM_recursion} on the $t$-th window is computed from the testing split $D_t^{\rm te}:= X_t^{\rm te} \cup \widetilde X_t^{\rm te} = \{ x_i \}_{i=1}^{2m'}$, $m' = (1-\alpha)w$ as follows,  where the binary labels $y_i$ are $0$ for reference and $1$ for arrival samples as before,
\begin{equation}\label{eq:def-etat-theory}
\eta_t 
= \frac{1}{m'}\sum_{i=1}^{2 m'} \left[y_i g_{\hat{\theta}_t}(x_i) - (1-y_i) g_{\hat{\theta}_t}(x_i)\right],
\end{equation}
%
which can be equivalently written as 
\begin{equation}\label{eta_def_1}
    {\eta}_t = \int_{\calX} g_{ \hat{\theta}_t }(x) ( \hat{f}_{1,t}^{\rm te} (x)- \hat{f}_{0,t}^{\rm te}(x))dx,
\end{equation}
where 
the empirical distributions of the test split of the $t$-th window from the arrival and reference samples are, respectively, 
\[\hat{f}_{1,t}^{\rm te} = \frac{1}{ m'} \sum_{x_i \in X_t^{\rm te} }\delta_{x_i},\quad \hat{f}_{0,t}^{\rm te} = \frac{1}{ m'} \sum_{x_i \in \widetilde X_t^{\rm te} }\delta_{x_i},\]  $\delta_{x}$ denoting the Dirac function. From \eqref{eta_def_1}, it can be seen that the increment depends on the training data through $\hat \theta_t$, and depends on test data through $\hat{f}_{1,t}^{\rm te} (x)- \hat{f}_{0,t}^{\rm te}(x)$.

As an initial step, we examine the drift term $\eta_t$'s necessary property so that the NN-CUSUM can detect a change. Suppose the test is done using the same deterministic function $g_\theta(x)$ instead of $g_{ \hat{\theta}}(x)$, it can be easily shown that 
\[\mathbb E_0 [\eta_t] = 0, \quad \mathbb E_1 [\eta_t] = \int g_\theta(x)[f_1(x) - f_0(x)] dx. \] Thus, if the test function $g_\theta(x)$ satisfies following {\it regularity condition} 
\begin{equation}
 \int g_\theta(x)[f_1(x) - f_0(x)] dx > 0,
\label{g_def}
\end{equation}
we have $0=\mathbb E_0[\eta_t] <  \mathbb E_1[\eta_t]$. So as long we can choose the drift term in between, $\mathbb E_0[\eta_t] < D < \mathbb E_1[\eta_t]$, we can have the desired CUSUM property: the expected increment is positive after the change, and negative before the change. 
Many types of neural networks can be employed to satisfy \eqref{g_def}, particularly the so-called neural divergence analysis. Existence and (ideal) estimation of $g_{\theta}$ represented by a neural network 
has been shown in the study of ``neural distance" \cite{zhang2018on} and neural estimation of divergences \cite{sreekumar2022neural}.
Theoretically, many $g_\theta$ can satisfy the regularity condition \eqref{g_def}, and it is a different question what $g_\theta$ will be found by neural network training.
In addition, the efficiency of change-point detection will also depend on the variance of $g_\theta$ beyond a positive bias.



In practice, the parameter $\hat \theta$ is estimated, and thus, it is also random and depends on the training batch. Below, we present a finite sample analysis. We will examine the {\it regularity condition} \eqref{g_def}. The analysis will take two steps: (i) When the trained model parameter $\hat\theta_t$ is held fixed (given the training data), relate the sample version $\eta_t$ 
to the population version $\mathbb E_1[\eta_t|\hat \theta_t] =\int g_{\hat \theta_t}(x) [f_1(x) - f_0(x)]dx$, in Section \ref{sec:test}; (ii) show that  $\mathbb E_1[\eta_t|\hat \theta_t]$ can be related to the Maximum Mean Divergence (MMD) statistic between $f_0$ and $f_1$, with high probability 
over the trained parameter (randomized with respect to the training data) in Section \ref{sec:NTK}.



\subsection{Effect of test samples: Concentration over testing data}\label{sec:test}

We will consider the deviation of $ \eta_t $  \eqref{eta_def_1} from its expected value conditioning on  the trained parameter $\hat\theta_t$, namely,
\[
\E_i [ \eta_t |\hat{\theta}_t  ] 
= \int
g_{ \hat{\theta}_t }(x) ( {f}_{i} (x)- {f}_{0} (x))dx, 
\quad i = 0, 1,
\]
which is determined by $D_t^{\rm tr}$. 
Conditioning on training, $ \eta_t $  is an independent sum
and thus, the deviation can be bounded using standard concentration
and the bound should depend on the test sample size $(1-\alpha)w$.
This is proved in the following lemma.

\begin{lemma}[Concentration on test samples]\label{lemma:conc-test}
Suppose the trained neural network function $g_{\hat \theta}(x)$ belongs to a uniformly bounded family $\calG$,  such that
$C: = \sup_{g \in \calG} \sup_{x \in \calX} |g(x)|  < \infty$, then  for any $\lambda_1 >0$, 
\[
\begin{aligned}
    &\mathbb{P}_{i} \left[ \eta_t - \E_i [ \eta_t |\hat{\theta}_t  ]  >  \frac{  2C \lambda_1 }{\sqrt{{(1-\alpha)w}}}  
\mid \hat{\theta}_t \right]\le e^{ - {\lambda_1^2}/{2}}, \\
&\mathbb{P}_{i} \left[  \eta_t - \E_i [ \eta_t |\hat{\theta}_t  ]  <- \frac{  2C \lambda_1 }{\sqrt{{(1-\alpha)w}}}  
\mid \hat{\theta}_t \right] 
\le e^{ - {\lambda_1^2}/{2}},
\end{aligned}
\]
for $i=0,1$.
The probability $\mathbb{P}_{i}$ is over the randomness of the testing split on $t$-th window.
\end{lemma}

\subsection{Effect of training: Analysis by Neural tangent kernel (NTK) MMD}\label{sec:NTK}


Below, we consider an approximation of $\eta_t$ in \eqref{eta_def_1} under the MMD loss, called $\eta_t^{\rm NTK}$. This is useful for subsequent analysis and can be linked to the neural tangent kernel (NTK) analysis \cite{cheng2021neural} in understanding neural network training dynamics. Extending NTK beyond MMD loss is an active research area. 

By definition, $\E_0 [ \eta_t |\hat{\theta}_t  ]   =0$ always.
To show that $\E_1 [ \eta_t |\hat{\theta}_t  ]  > 0$, we need to analyze the training guarantee. 
For training using the MMD loss \eqref{eq:def-l-mmd}, 
the guarantee holds with high probability as shown in \eqref{eq:E1-etat-cond-NTK} by Lemma \ref{lemma:NTK-etat-bound}. For a fixed window $(t-w, t]$, the training of MMD loss \eqref{eq:def-l-mmd} is conducted by one-pass stochastic gradient decent SGD minimization of the $2m$ samples with a small learning rate $\gamma > 0$ (allowed by numerical precision).
We denote the trained parameter as $\hat{\theta} = \hat{\theta}_t$ and the random initial parameter as $\theta_0$. 
The increment $\eta_t $ is computed by specifying $g_{\hat{\theta}_t}$ as 
\begin{equation}
    g_{\hat{\theta}_t }(x) = \frac{1}{\gamma} (  \varphi (x, {\hat{\theta}_t}) - \varphi  ( x, {{\theta_0}} ) ).
\end{equation}
By the NTK kernel approximation under technical assumptions of the neural network function class \cite{cheng2021neural},
we know that for every $x$ inside the data domain $\calX$,
\begin{equation}\label{eq:ntk-approrx-gtheta}
|g_{\hat{\theta}_t}(x) 
- \hat{g}^{\rm NTK}(x) |
\le C_K \gamma,
\quad 
\hat{g}^{\rm NTK}(x):=
\int_\calX K_0(x, x') (\hat{f}_{1,t}^{\rm tr} (x') - \hat{f}_{0,t}^{\rm tr} (x') )  dx', 
\end{equation}
where 
the empirical distributions of the training split are defined as
\[
\hat{f}_{1,t}^{\rm tr} = \frac{1}{ m} \sum_{x_i \in X_t^{\rm tr} }\delta_{x_i},
\quad 
\hat{f}_{0,t}^{\rm tr} = \frac{1}{ m} \sum_{x_i \in \widetilde X_t^{\rm tr} }\delta_{x_i},\]
and the zero-time finite width NTK:
\[
K_0(x,x') := \langle \partial_\theta \varphi (x,\theta_0),  \partial_\theta \varphi (x',\theta_0) \rangle
\]
The constant $C_K$ in \eqref{eq:ntk-approrx-gtheta} depends on the theoretical boundedness of the network function and its derivatives on $\calX$ over short-time training, based on Proposition 2.1 and Appendix A.2 in \cite{cheng2021neural}.

The approximation \eqref{eq:ntk-approrx-gtheta} gives that
the increment ${\eta}_{t}$ defined in \eqref{eta_def_1} can be approximated  by  ${\eta}_{t}^{\rm NTK} $ defined as 
\begin{equation}\label{eq:def-eta-t-exactNTK}
{\eta}_{t}^{\rm NTK} 
:= \iint K_0(x, x') (\hat{f}_{1,t}^{\rm tr} (x')- \hat{f}_{0,t}^{\rm tr} (x')) 
	( \hat{f}_{1,t}^{\rm te} (x)- \hat{f}_{0,t}^{\rm te}(x))  dx'  dx.
\end{equation} 
Specifically, 
for the expected increment condition on training, 
\eqref{eq:ntk-approrx-gtheta} implies that 
\begin{align}
\E_{1} [{\eta}_{t}| \hat \theta_t  ]
& = \int_{\calX} g_{\hat\theta_t}(x) ( f_{1}(x)- {f}_{0}(x))dx \nonumber \\
& = \int_{\calX} \hat{g}^{\rm NTK}(x) ( f_{1}(x)- {f}_{0}(x))dx + \text{e}_t \nonumber \\
& = \E_{1} [{\eta}_{t}^{\rm NTK}| D_t^{\rm tr}  ]   + \text{e}_t ,
\quad 
| \text{e}_t | \le  2 C_K \gamma.
\label{eq:E1-cond-E1-ntk-approx}
\end{align}

Meanwhile, note that the full expectation $\E_{1} [{\eta}_{t}^{\rm NTK}]$ equals the squared population kernel MMD between $f_1$ and $f_0$ \cite{gretton2012kernel} with respect to the kernel $K_0$
\begin{equation}
 \E_{1} [{\eta}_{t}^{\rm NTK}]={\rm MMD}^2( f_1, f_0)
 :=\iint K_0(x, x') ({f}_{1} (x')- {f}_{0} (x')) 
	( {f}_{1} (x)- {f}_{0} (x))  dx'  dx.
\end{equation}
By definition, 
$ {\rm MMD}^2( f_0, f_0) =0$.
We assume $ {\rm MMD}^2( f_1, f_0) > 0$ when $f_1 \neq f_0$, 
that is, the NTK kernel $K_0$ is discriminative enough to produce a positive $ {\rm MMD}^2( f_1, f_0)$ post change. Combining above, we have the positive expected drift
\begin{equation}\label{positive_exp}
\mathbb E_1 [\eta_t]  =  \E_{1}  [\E_{1} [{\eta}_{t}| \hat \theta_t  ]] =  {\rm MMD}^2( f_1, f_0)  + \mathbb E_1 [\text{e}_t] \geq {\rm MMD}^2( f_1, f_0) - 2C_K \gamma.
\end{equation}
Thus, $\mathbb E_1 [\eta_t] >0$ when $\gamma$ is sufficiently small related to the population squared MMD.

In addition, the following lemma provides the guarantee of the positivity of  $\E_1 [\eta_t^{\rm NTK} | D_t^{\rm tr}]$ at sufficiently large finite training sample size $ \alpha w$.

\begin{lemma}[Guarantee under MMD loss]
\label{lemma:NTK-etat-bound}
Suppose $\sup_{x,x' \in \calX} K_0(x,x') \le B$, $B$ being a positive constant.
For any $\lambda_2 > 0$, after the change,  w.p. $\geq 1-e^{-\lambda_2^2/2}$,
\begin{equation}\label{MMD_lemma}
 \E_1 [\eta_t^{\rm NTK} |  D_t^{\rm tr}]
 \ge {\rm MMD}^2( f_1, f_0) -   \frac{\lambda_2 B}{ \sqrt{\alpha w}}.  
\end{equation}
The good event 
is over the randomness of training data $ D_t^{\rm tr}$ on the $t$-th window. 
\end{lemma}

Combining \eqref{MMD_lemma} with \eqref{eq:E1-cond-E1-ntk-approx}, this suggests that 
with probability $\ge 1-e^{-\lambda_2^2/2}$ with respect to the randomness of training data $ D_t^{\rm tr}$ on the $t$-th window
\begin{equation}\label{eq:E1-etat-cond-NTK}
\E_{1} [{\eta}_{t}| \hat \theta_t  ]
\ge {\rm MMD}^2(f_1, f_0) 
- \left(\frac{\lambda_2 B}{ \sqrt{\alpha w}}
+ 2 C_K \gamma 
\right).
\end{equation}
Combining with the concentration over test samples in Lemma \ref{lemma:conc-test}, we have that with probability $\geq 1-e^{-\lambda_1^2/2} - e^{-\lambda_2^2/2}$,
\begin{equation}\label{increment_high_prob}
\eta_t \geq {\rm MMD}^2(f_1, f_0) -\left( 
\frac{2C\lambda_1}{\sqrt{(1-\alpha)w}} +
\frac{ B\lambda_2}{ \sqrt{\alpha w}}
  + 2 C_K \gamma
\right).
\end{equation}
Thus, when the ``signal'' ${\rm MMD}^2(f_1, f_0)$ is sufficiently large, 
$w$ is sufficiently large,
and $\gamma$ is controlled to be small, we can ensure the increment is positive with high probability. When choosing the drift-term $D$ to be sufficiently small, we can ensure the NN-CUSUM increment $\eta_t - D$ is still positive with high probability.

\subsection{EDD and ARL performance}\label{sec:ARL and EDD}

Now, we examine the detection property. Note that, unlike classic CUSUM analysis, where the increments are log-likelihood ratio based (e.g., exact or with plug-in parameter estimators); here the increments are not directly related to likelihood ratio. So, we cannot invoke the classic results (e.g., Section 6 in \cite{siegmund1985sequential}). Yet, we can still use Wald's identity-like analysis to analyze how quickly the NN-CUSUM procedure can detect a change.

For sequential change-point detection, two commonly used performance metrics are: (i) the average run length (ARL), representing the expected value of the stopping time in the absence of any changes, and (ii) the expected detection delay (EDD). Below, we consider EDD when a change occurs immediately at the beginning, rather than the worst-case detection delay, to simplify the analysis. The EDD analysis will shed light on algorithm design and detection power. We leave the analysis of theoretical ARL to future work; the choice of the threshold to control ARL in practice is explained in Section \ref{sec:threshold}. 



\subsubsection{i.i.d. increments}\label{iid_eta}

First, consider a simplified setting: in the training process, the training and testing windows are non-overlapping from the current time to the next, which can happen when we choose stride $s=w$. Moreover, assume the parameters $\theta_t$ are trained with new initialization, so $\theta_t$ are also not independent across $t$. In this setting, we can assume $\eta_t$ are i.i.d. Then we can obtain a simple estimate for the expected detection delay. We would like to remark that the following analysis can be used for general non-parametric CUSUM with increments with necessary properties.

We first state the classic Wald's identity, to be self-contained.
\begin{lemma}[Wald's identity]\label{Walds}
Let $X_n$, $n =1, 2, \ldots$ be an infinite sequence of real-valued random variables, and let $N$ be a nonnegative integer-valued random variable. Assume that (i) $X_n$ are all integrable (finite-mean); (ii) $\mathbb E[X_n \mathbb I\{N\geq n\}]=\mathbb E[X_n] \mathbb P(N\geq n)$ for all $n$, and (iii) the infinite series satisfies
\[
\sum_{n=1}^\infty \mathbb E[|X_n|\mathbb I\{N\geq n\}] <\infty.
\]
Then, the random sums below are integrable, and 
$
\mathbb E\left[\sum_{n=1}^N X_n\right] = \mathbb E\left[\sum_{n=1}^N \mathbb E[X_n]\right].
$
If, in addition, (iv) $\mathbb E X_n = \mathbb E X_1$, for all $n$, and (v) $\mathbb E N< \infty$, 
then $\mathbb E\left[\sum_{n=1}^N X_n\right] = \mathbb E[N] \mathbb E[X_1].$
\end{lemma}

The CUSUM recursion is not a simple random walk. For notational simplicity, define $\tilde \eta_t = \eta_t - D$. Also define 
\begin{equation}
    \tilde S_t = \sum_{n=1}^t \tilde \eta_t, \quad t = 1, 2, \ldots, \quad \tilde S_0 = 0.
\end{equation}
To derive the EDD bound for $\tau$, we can show using standard derivation, the detection statistic for CUSUM \eqref{eq:NNCUSUM_recursion} is given by 
\begin{equation}
    S_t = \max_{1\leq k< t}\sum_{n=k}^t \tilde \eta_n = \tilde S_t - \min_{0\leq n\leq t} \tilde S_n. \label{S_alternative}
\end{equation}

Similar to Stein's lemma that was originally derived for the sequential likelihood ratio test (see, e.g., Proposition 2.19 in \cite{siegmund1985sequential}), we can show the following (which can hold both under $f_0$ and $f_1$):
\begin{lemma}[Finite expected stopping time for general CUSUM with i.i.d. increments]\label{CUSUM_finite}
Let $X_1, X_2, \ldots$ be i.i.d. 
Consider the CUSUM recursion $S_n = \max\{S_{n-1}+X_n, 0\}$, $S_0=0$. Define $N = \inf\{n: S_n \geq b\}$. Suppose for some small $\delta \in (0, 1)$, $\mathbb P\{X_1\geq \delta\}\geq \delta$. Then $\mathbb E\{N\}<\infty$.
\end{lemma}

The precise characterization of the EDD of general CUSUM can be done by generalizing the technique for likelihood ratios in (Section II.6, \cite{siegmund1985sequential}) using non-linear renewal theory and passing the limit of $b\rightarrow \infty$. 
Below, we present a simpler approach to derive an upper bound for the EDD of NN-CUSUM.

\begin{theorem}[EDD bound for NN-CUSUM, i.i.d. increments]
\label{main_thm}
Assume $\mathbb E_1 [|\eta_t|] < \infty$, $\eta_t$ are i.i.d., and $w=  O(1/({\rm MMD}^2(f_1, f_0)^2)$, we have  
    \begin{equation}\label{EDD_approx}
          \mathbb E_1 [\tau]  \leq \frac{b}{{\rm MMD}^2( f_1, f_0) - 2C_K \gamma-D} + O(1),
     \end{equation}
\end{theorem}
\begin{remark}
In \eqref{EDD_approx}, $O(1)$ is due expected overshoot $\mathbb E_1
[\tilde S_{\tau} - D - b]$, which is usually on the order of the expected increment; an analysis can be found in, e.g., \cite{siegmund1985sequential,xie2022window}.
\end{remark}

Comparing the upper bound for EDD \eqref{EDD_bound} with likelihood-ratio based CUSUM, which is $b/{\rm KL}(f_1||f_0)$,  where ${\rm KL}$ denotes the Kullback-Leibler divergence between the two distributions, we can see that due to the choice of neural network training loss and resulted from NTK, we have a looser upper bound compared to the known good bound for likelihood-ratio based CUSUM, as expected. If we choose logistic loss as the training loss, in theory, we could achieve an EDD upper bound $b/{\rm KL}(f_1||f_0)$; however, in practice, due to the highly non-linearity of logistic loss, we cannot generalize the NTK analysis therein, and in practice, it is known to be hard to ensure the trained neural network to converge to the log-likelihood function.

\begin{remark}[ARL approximation for NN-CUSUM, i.i.d. increments] \label{ARL_CUSUM} It can be shown that when there is no change, under $f_0$, $\tau(b)$ in \eqref{eq:NNCUSUM} is asymptotically exponentially distributed. Here, we explicitly denote the dependence of the stopping time on threshold $b$. The argument can be based on a Poisson approximation argument similar to Lemma D.1 in \cite{li2019scan} and  Theorem 1 in \cite{yakir2009multi}, or more using the change-of-measure technique of Lemma 2 in \cite{wei2022online} when $\eta_t$ are i.i.d.. Intuitively, for large value of $b$, we expect the probability $\mathbb P_0\{\tau(b) > t\}$ to be close to  $e^{-\lambda t}$ at least when $t$ is also large. 
The approximation is accurate even in the non-asymptotic regime, as shown by numerical examples in Section \ref{sec:ARL}. Such asymptotic exponential distribution property of $\tau$ is helpful to simplify the ARL simulation for threshold calibration, as explained in Section \ref{sec:threshold}.
\end{remark}

\subsubsection{\texorpdfstring{$m$}{m}-dependent stationary process}

In most scenarios, we may have dependent $\eta_t$. This may happen when the stride is shorter than the window length $w$, and thus, from $t$ to $t+1$ (the next moment), there is a significant overlap in the training and testing stack data from $t$ to $t +1$. However, due to the design of the algorithm, after $w/s$ steps, the training and testing stacks at time $t+w/s$ will have completely fresh data that is non-overlapping with $t$. If the neural network is training using random initialization using only the data from the current training stack, we can assume that $\eta_t$ is $m$-dependent, for $m = w/s$, meaning that $\eta_t$ is independent of past $\eta_{t-s}$ if $s\geq w/s$.

Thus, we need Wald's lemma for dependent but finite memory sequences to carry out the EDD analysis similar to the previous section. Such results have been developed before in \cite{moustakides1999extension}. Below, to be self-contained, we derive a version of the result suitable for our scenario, following a similar proof strategy as in \cite{xie2022window}.

%
\begin{lemma}[Generalized Wald's identity for $m$-dependent sequence]\label{generalized_walds}
Consider a stationary sequence of random variable $X_n$, $n = 1, 2, \ldots$ with finite and positive mean $\mathbb E X_n > 0$, furthermore, the increments $X_n$ are $m$-dependent, i.e., $X_n$ is independent of $\mathcal F_{n-s}$ for all $s \geq m$.
Consider the stopping time $N = \inf \{t>m: \sum_{n=m+1}^{t} X_n > b\}.$ If $E[N]<\infty$, then, we have
\[
\mathbb E [N] = \frac{b}{\mathbb E[X_1]} + m + O(1).
\]
where the $\mathcal O(1)$ comes from the overshoot of random sum over the threshold $b$ when stopping at time $N$.
\end{lemma}
%



\begin{theorem}[EDD upper bound under $m$-dependent stationary process]\label{EDD_dependent} 
If $\eta_t, t = 1, 2 \ldots$ forms a stationary process, and $\mathbb E_1[\tau] < \infty$, we have 
\begin{equation}\label{m_dependence}
    \mathbb E_1[\tau] \leq  \frac{b}{{\rm MMD}^2( f_1, f_0) - 2C_K \gamma-D} + w + O(1).
\end{equation}
\end{theorem}
The proof of Theorem \ref{EDD_dependent} follows a similar argument as those for Theorem \ref{main_thm}, and uses Lemma \ref{generalized_walds}, which requires $\eta_t$ to be stationary. Since our data is i.i.d., and the dependence is introduced by overlapping training and testing stack, it would be reasonable to believe $\eta_t$ is stationary (note that we can allow $\eta_t$ to be dependent up to memory length $m$.) We assume $\mathbb E_1[\tau] < \infty$: the verification needs to generalize the arguments in Lemma \ref{CUSUM_finite} for $m$-dependent sequence is technical so we decide to omit; however, we observe most of the time in practice after the change the detection will stop  in a reasonable time. In comparing \eqref{m_dependence} with \eqref{EDD_bound}, we have an extra term $w$ introduced due to the memory of $\eta_t$ due to overlapping windows.

\begin{remark}[Effect of window length $w$]\label{rmk:effect_w}
From the analysis so far, we can see the choice of the window length faces a trade-off. On the one hand, from \eqref{increment_high_prob}, we see that to ensure $\eta_t$ is close to desired ``signal'' ${\rm MMD}^2( f_1, f_0)$ in high probability, we need $w$ to be sufficiently large, to control the terms ${2C\lambda_1}/{\sqrt{(1-\alpha)w}}$ (due to testing data randomness)
${ B\lambda_2}/{ \sqrt{\alpha w}}$ (due to training data randomness) to be small. Conversely, a larger $w$ contributes linearly to a larger upper bound for EDD in \eqref{m_dependence}. So there may exist some optimal choice of $w$ that is neither too small, nor too large; we observe such a phenomenom in experiment, as shown in Section \ref{sec:opt_window}.
\end{remark}

For future work, we can extend Remark \ref{ARL_CUSUM} for ARL approximation, from i.i.d. increments to $m$-dependent case, by using the techniques similar to \cite{li2019scan} and the generalized Poisson approximation to allow for dependence \cite{arratia1989two}. 


\section{Experiments}
\label{sec:experiments}

To examine the performance of NN-CUSUM, we compare the proposed method with various baselines, including NN-based and classic methods, on simulated datasets and also real datasets.
Before the comparison, we verify the exponential distribution of $\tau$ under $f_0$ and introduce some details of the algorithm, 
including 
how to calibrate the threshold to control the false alarms (detect a change when there is none)
and numerical studies to investigate the properties of the increments and the existence of an optimal window.


\begin{figure}[hb!]
\centering 
\includegraphics[width=0.4\linewidth]{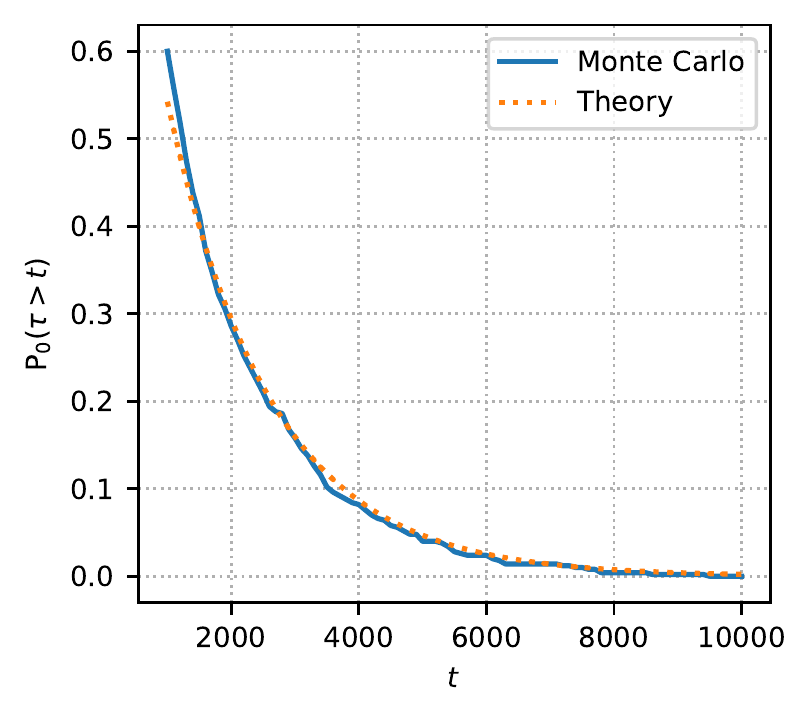}
\caption{The tail probability $\mathbb P_0\{\tau>t\}$ versus $t$. The data consists of 500 sequences with lengths of 40000, generated from i.i.d. Gaussian distribution with a dimension of 100. On this data, we perform NN-CUSUM to get $\eta_t$s. The threshold $b = 1$. The solid curve shows numerical values from 500 Monte Carlo repetitions. The dashed curve represents the theoretical values, $e^{-\lambda t}$ with $\lambda = 6\times 10^{-4}$, which is fitted by non-linear regression on the numerical values.
}
\label{fig:Exponential_stoppingtime}
\end{figure}
\subsection{Verifying exponential distribution of \texorpdfstring{$\tau$}{tau} under \texorpdfstring{$f_0$}{f0}}\label{sec:ARL}

Now we verify Remark~\ref{ARL_CUSUM}  that the asymptotic distribution of $\tau$ under $f_0$ is exponential. 

\vspace{3pt}
\noindent{\it Setup.} 
In this example, the neural network architecture of NN-CUSUM consists of one fully connected hidden layer of width 64 with a rectified linear unit (ReLU) activation. The last layer is a one-dimensional, fully-connected layer without activation. The training batch size is 100. The stride of the sliding window is 10. We use Adam~\cite{kingma2014adam} for training under the logistic loss function with a learning rate of $10^{-3}$ without scheduling.
We set window size $w=200$ with a split ratio $\alpha = 0.5$. We generate $500$ sequences with a total length $40000$ consisting of i.i.d. standard Gaussian random variables in $\mathbb R^{100}$. This example will have no change point since we study the tail probability of $\tau$ under $f_0$. We set the threshold $b=1$ in \eqref{eq:NNCUSUM}.

\vspace{3pt}
\noindent{\it Results.} In Figure~\ref{fig:Exponential_stoppingtime}, the solid curve shows the empirical tail probability of $\tau$ from Monte Carlo trials, which we fit with the curve $\mathbb P\{\tau > t\} = e^{-\lambda t}$ using non-linear regression and to find $\lambda = 6\times 10^{-4}$. Note from Figure~\ref{fig:Exponential_stoppingtime} that this leads to a good match between $e^{-\lambda t}$ and the empirical tail probabilities.

\subsection{Calibration of threshold \texorpdfstring{$b$}{b}}\label{sec:threshold}

\begin{figure}[b!]
\centering 
\includegraphics[width=0.4\linewidth]{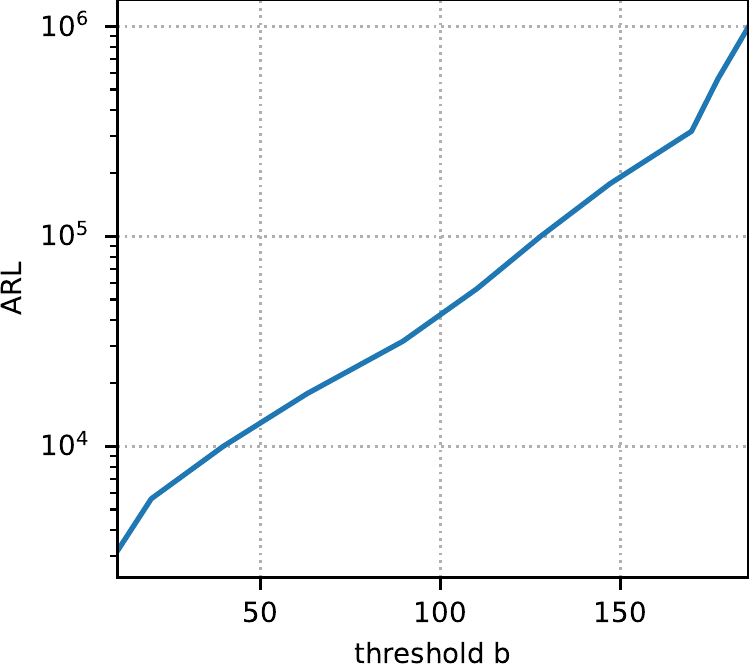}
\caption{Numerically estimate ARL versus threshold $b$; data is i.i.d. following a Gaussian mixture model $x_i\sim 1/2 \mathcal N(2 {\bf 1}, I_d) + 1/2 \mathcal N(-2 {\bf 1}, I_d)$ where ${\bf 1}$ denotes an all-one vector, and $I_d$ is a $d$-by-$d$ identity matrix (corresponds to the $f_0$ of the fourth example in Table~\ref{tab:simulated-distributions}). The experiment consists of 400 sequences of length 500.
}
\label{fig:ARL_vs_b}
\end{figure}

It is computationally expensive to use direct simulation of $\tau$ to estimate its mean $\mathbb E_0[\tau]$. With each fixed threshold $b$, a large number of repetitions is required, and each repetition takes a long time because we are typically interested in tuning the threshold $b$ to obtain a large ARL when there is no change.
Thus, we use a similar technique as in \cite{xie2013sequential} to speed up the simulation to estimate ARL. Recall from Remark \ref{ARL_CUSUM}, when the data are i.i.d. $f_0$ (i.e., before the change-point), the stopping time $\tau$ is asymptotically exponentially distributed for some $\lambda > 0$.
Then we can estimate for $\bbP_0[\tau \geq T]\approx \exp\{-\lambda T\}$, to evaluate an estimate of the rate $\hat \lambda$. On the other hand, for the exponential random variable, its mean is equal to the inverse of the rate, and then ARL can be approximated by $\mathbb E_0[\tau] \approx 1/\lambda$. This way, we can convert simulating $\tau$ (which usually takes long runs because we are interested in large ARL when there is no change) into simulating fixed length $T$ sequences and estimating the probability of such sequences exceeding a threshold $b$. 

Figure~\ref{fig:ARL_vs_b} shows ARLs versus the thresholds $b$ on an example consisting of 400 pre-change sequences from GMM and verified that the logarithm of ARL is approximately proportional to threshold $b$, and common property for CUSUM procedures \cite{xie2021sequential}. 
\begin{table}[b!]
    \caption{Mean of pre- and post-change $\eta_t$ in the Gaussian sparse mean shift example. The standard deviations are in parentheses.}
    \label{tab:eta-pre-post}
    \centering
\begin{scriptsize}
\begin{tabular}{ccc}
\toprule
$w$  & Pre-change & Post-change  \\
\midrule
20 & 0.001 (0.1441) & 0.411 (0.4607) \\[1pt]
60 & 0.003 (0.1685) & 0.711 (0.3322) \\[1pt]
100 & -0.026 (0.2092) & 1.337 (0.3315) \\[1pt]
200 & 0.022 (0.3676) & 1.621 (0.5112) \\
\bottomrule
\end{tabular}
\end{scriptsize}
\end{table}
\begin{figure}[b!]
\centering 
\begin{subfigure}[h]{0.24\linewidth}
\includegraphics[width=\linewidth]{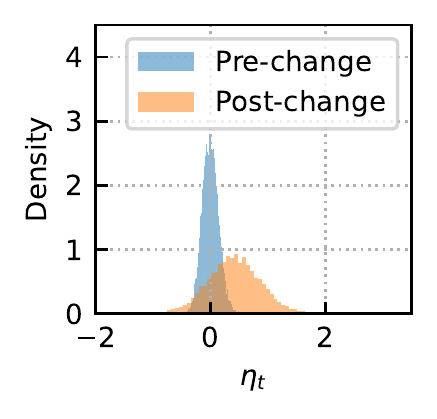}
\caption{$w=20$}
\end{subfigure}
\begin{subfigure}[h]{0.24\linewidth}
\includegraphics[width=\linewidth]{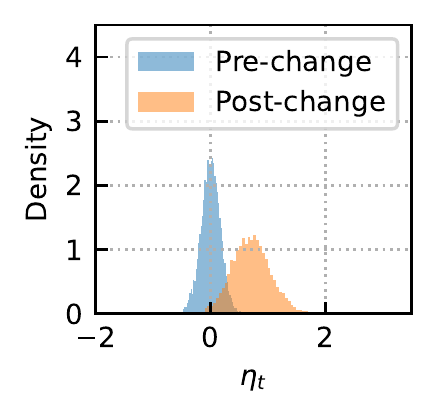}
\caption{$w=60$}
\end{subfigure}
\begin{subfigure}[h]{0.24\linewidth}
\includegraphics[width=\linewidth]{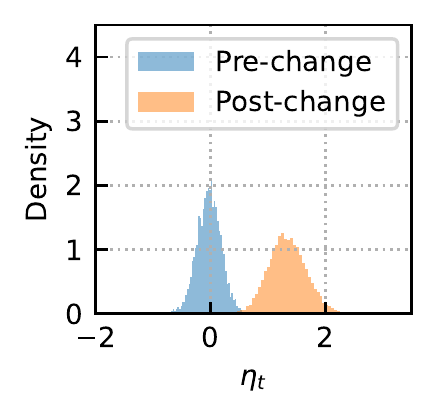}
\caption{$w=100$}
\end{subfigure}
\begin{subfigure}[h]{0.24\linewidth}
\includegraphics[width=\linewidth]{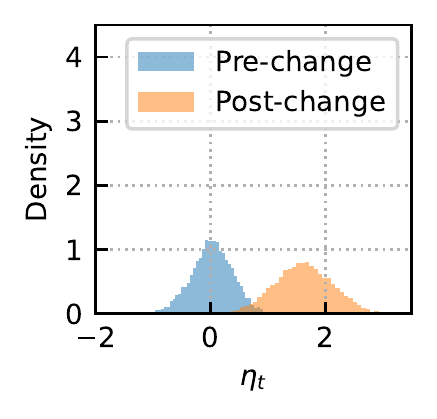}
\caption{$w=200$}
\end{subfigure}
\caption{Pre- and post-change distributions of $\eta_t$ in the Gaussian sparse mean shift example.}
\label{fig:eta-t_distribution_mean_shift}
\end{figure}

\subsection{Properties of increment \texorpdfstring{$\eta_t$}{etat}}
\label{subsec:num-just-incre-loglike}

We empirically verify that the training of neural networks will produce nearly zero pre-change $\eta_t$ and significantly positive post-change $\eta_t$.

\vspace{3pt}
\noindent{\it Setup.}
The neural network setup of NN-CUSUM is similar to Section~\ref{sec:ARL}, despite the width of the hidden layer 512 and the batch size of 10. 
The neural network is trained on the Gaussian sparse mean shift example, namely the first example in Table~\ref{tab:simulated-distributions} where we set $\mu_q = (\delta, \delta/2, \delta/3, 0, \cdots, 0)^\top$ and  $\delta=0.8$. 

We fix the split ratio to be $\alpha=0.5$, whereas the window sizes slide in $\{20,60,100,200\}$. The data is generated as a sufficiently long sequence with a length $2\times 10^5$ to ensure the stability of the sample average means on $\eta_t$. The mean shift happens in the middle, namely at $1\times 10^5$. For the study on log-likelihood ratio, we set $w=300$ and $\alpha = 1/3$ in training. The data is generated with a change point at 5000. The time horizon varies in $\{5000, 5080, 5160\}$. At a given time horizon, we generate 1000 testing points, in which half comes from $f_0$ and another half comes from $f_1$, and we compute both the true and learned log-likelihood ratio on testing data. 

\vspace{3pt}
\noindent{\it Results.}
Table~\ref{tab:eta-pre-post} shows the temporal means of pre- and post-change $\eta_t$. 
The post-change expected $\eta_t$ becomes positive with $w=20$, and the magnitude is more significant as the window size increases. 
Figure~\ref{fig:eta-t_distribution_mean_shift} visualizes the empirical distribution of the pre- and post-change $\eta_t$ under different $w$,
 showing that as the sizes of training and testing windows both increase, the mean shift in  $\eta_t$ becomes more evident, representing a possibly faster detection (consistent with Table~\ref{tab:eta-pre-post}).  
In this example, the NN-CUSUM procedure obtains a positive expected $\eta_t$
at $w$ as small as 20,
which, according to Section \ref{sec:theory justification}, suggests that the method can detect the change point.
There is a trade-off in choosing window size $w$ because larger $w$ may delay the detection, though it results in a larger drift in $\eta_t$.
%



\begin{figure}[t!]
\centering 
\begin{subfigure}[h]{0.4\linewidth}
\includegraphics[width=\linewidth]{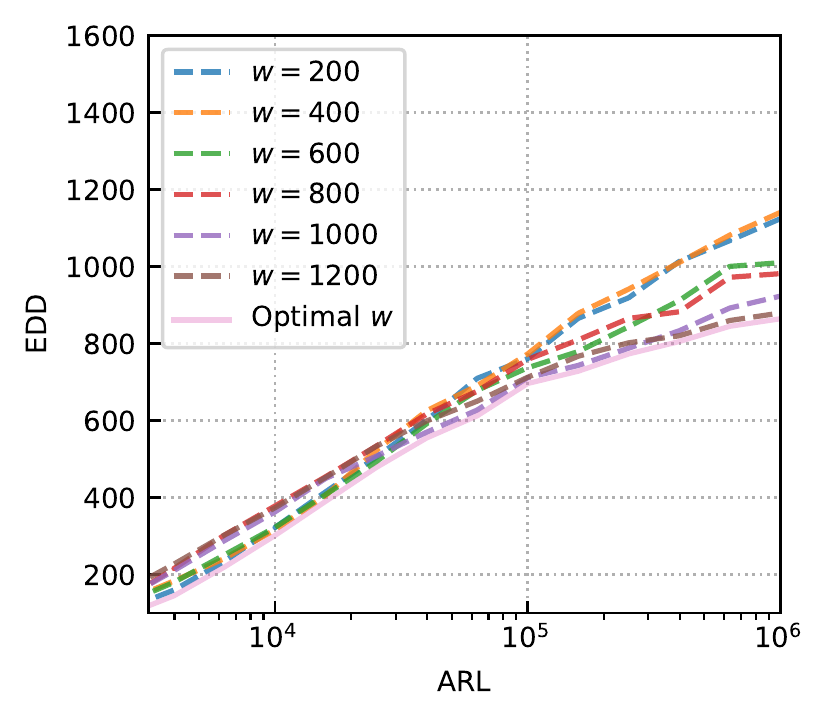}
\caption{Minimal EDD}
\label{fig:minimal_EDD}
\end{subfigure}
\hspace{+30pt}
\begin{subfigure}[h]{0.4\linewidth}
\includegraphics[width=\linewidth]{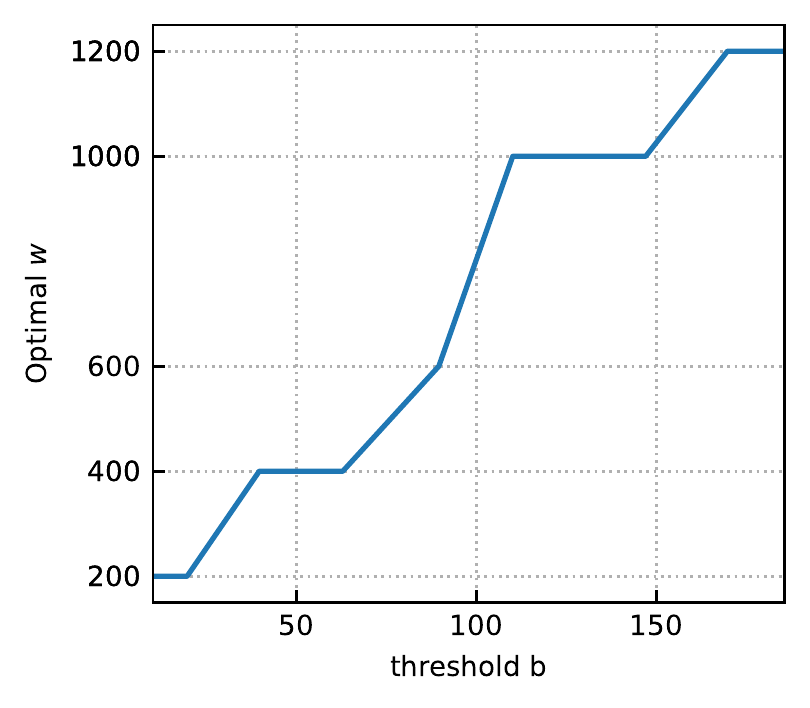}
\caption{Optimal $w$ versus $b$}
\label{fig:optimal_w_vs_b}
\end{subfigure}

\caption{Effect of window length $w$: (a) Minimal ARL versus EDD over window sizes from 200 to 1200 by 200. (b) Optimal window size $w$ versus threshold $b$. 
The experiment is conducted on the GMM component shift example in Section~\ref{sec:opt_window}. 
}
\label{fig:num_just}
\end{figure}

\subsection{Optimal window size}\label{sec:opt_window}

We empirically investigate the trade-off in choosing the window size $w$ as discussed in Remark \ref{rmk:effect_w}. 

\vspace{3pt}
\noindent{\it Setup.}
The neural network structure, batch size, and stride are the same as in Section~\ref{subsec:num-just-incre-loglike}. The split ratio $\alpha$ takes value in $\{0.1,0.2,0.4,0.8\}$, and the window size varies in $\{200, 400, \cdots, 1200\}$. The example here is the GMM with a component shift, defined in the fourth example in Table~\ref{tab:simulated-distributions}. To be concrete, we let $f_0\sim 1/2 \mathcal N(2 {\bf 1}, I_d) + 1/2 \mathcal N(-2 {\bf 1}, I_d) $ and $f_{1} {\sim} 1/3 \mathcal N(2 {\bf 1}, I_d) + 1/3 \mathcal N(-2 {\bf 1}, I_d) + 1/3 \mathcal N\left((13/5){\bf 1}, 0.8I_d+0.2E\right)$ where ${\bf 1}$ is a $d$-dimensional all-one vector, $I_d$ is a $d$-by-$d$ identity matrix and $E$ is a $d$-by-$d$ all-one matrix.
Four hundred sequences are generated, with the lengths of each 10000 and the change point at the middle of 5000.

\vspace{3pt}
\noindent{\it Results.}
In Figure~\ref{fig:minimal_EDD}, the dotted curves are the lowest EDD over all $\alpha$'s for every fixed $w$. The optimal EDD achieves the lower envelope of EDDs over every pair of $w$ and $\alpha$. The optimal $w$, to which the lower envelope overlaps with the corresponding curve, increases with the growth of ARLs. Figure~\ref{fig:optimal_w_vs_b} justifies Remark~\ref{rmk:effect_w} and indicates that the optimal $w$ monotonously increases versus a growing threshold $b$. Figure~\ref{fig:ARL_vs_b} characterizes the monotone relationship between ARLs and $b$s. The combination of Figure~\ref{fig:optimal_w_vs_b} and~\ref{fig:ARL_vs_b} supports Figure~\ref{fig:minimal_EDD} in terms of the transitions of optimal $w$ against ARLs.

\begin{table}[t!]
    \centering
\caption{Ten examples in simulations: $\mu_q$ is the mean shift or the location shift, $D$ is a diagonal matrix, $E$ is an all-ones matrix, $\kappa$ is a shape parameter, and $\beta_i,\, i=0,1$ are scale parameters. In GMM, $\phi_{i,0}$ and $\phi_{i,1}$ are pre- and post-change mixture weights; then $\mu_{i,0}$, $\mu_{i,1}$ and $\Sigma_{i,1}$ are the means and covariances in corresponding components. In non-central Chi-square distributions, $\kappa_c$ is the degree-of-freedom, and $\lambda_i,\, i=0,1$ is the non-centrality. In Pareto distributions, $x_m$ is the lower bound of variables, and $b_{i},\,i=0,1$ are shape parameters. More details can be found in Appendix~\ref{app:additional_simulation}.}
\begin{scriptsize}
\label{tab:simulated-distributions}
    \begin{tabular}{llll}
        \toprule
        Index & Distribution & Pre-change & Post-change\\
        \midrule
        1 & Gaussian & $f_{0} {\sim}\calN(0, I_d)$ & $f_{1} {\sim}\calN(\mu_q, I_d)$ \\[2pt]
        2 & Gaussian & $f_{0} {\sim}\calN(0, I_d)$ & $f_{1} {\sim}\calN(0, I_d-D^2+DED)$\\[3pt]
        3 & Log Gaussian  & $f_{0} {\sim}\exp\{\calN(0, I_d)\}$ & $f_{1} {\sim}\exp\{\calN(0, 0.8I_d+0.2E)\}$\\[3pt]
        4 & GMM & $f_{0} {\sim} 1/2 \mathcal N(2 {\bf 1}, I_d) + 1/2 \mathcal N(-2 {\bf 1}, I_d)$ & $f_{1} {\sim} 1/3 \mathcal N(2 {\bf 1}, I_d) + 1/3 \mathcal N(-2 {\bf 1}, I_d) + 1/3 \mathcal N(\mu_q, \Sigma_q)$ \\[3pt]
        5 & Non-central Chi-square & $f_{0i} \stackrel{i.i.d.}{\sim} \chi^2(\kappa_c,\nu_0)$ & $f_{1i} \left\{
        \begin{array}{ll}
             \stackrel{i.i.d.}{\sim}\chi^2(\kappa_c,\nu_1), & i\in\{1,26,51,76\} \\
             \,\,\,\, = f_{0i}, & \text{otherwise}
        \end{array}\right.
            $ \\[6pt]
        6 & Pareto & $f_{0i} \stackrel{i.i.d.}{\sim} \text{Pareto}(x_m,b_0)$ & $f_{1i} \stackrel{i.i.d.}{\sim}\text{Pareto}(x_m,b_1)$\\[2pt]
        7 & Exponential & $f_{0i} \stackrel{i.i.d.}{\sim} \text{Exp}(\beta_0)$ & $f_{1i} \stackrel{i.i.d.}{\sim}\text{Exp}(\beta_1)+\mu_q$ \\[2pt]
        8 & Gamma & $f_{0i} \stackrel{i.i.d.}{\sim} \text{Gamma}(\kappa,\beta_0)$ & $f_{1i} \stackrel{i.i.d.}{\sim}\text{Gamma}(\kappa,\beta_1)+\mu_q$\\[2pt]
        9 & Weibull & $f_{0i} \stackrel{i.i.d.}{\sim} \text{Weibull}(\kappa,\beta_0)$ & $f_{1i} \stackrel{i.i.d.}{\sim}\text{Weibull}(\kappa,\beta_1)+\mu_q$\\[2pt]
        10 & Gompertz & $f_{0i} \stackrel{i.i.d.}{\sim} \text{Gompertz}(\kappa,\beta_0)$ & $f_{1i} \stackrel{i.i.d.}{\sim}\text{Gompertz}(\kappa,\beta_1)+\mu_q$ \\
        \bottomrule
    \end{tabular}
    \end{scriptsize}
\end{table}

\subsection{Comparison on simulated data}\label{subsec:numerical-comparison}

We perform a comprehensive numerical comparison of the proposed NN-CUSUM with baselines and related existing methods. 

\vspace{.1in}
\noindent{\it Setup.} In simulation comparisons, the neural network architecture of NN-CUSUM is the same as the example in Section~\ref{sec:ARL}. The window size is $w=200$ with a split ratio $\alpha = 0.5$. We generate the sequences with a total length $5500$ and the change-point located at $k=500$, namely the pre-change length $500$. 

In our experiments, we calibrate the threshold to achieve a constant ARL and the Type-I error (for real-data experiments), which is the probability that the procedure detects a change before the true change point. Using the ARL calibration in Section \ref{sec:threshold}, this choice, along with the fixed $\ARL=5000$ later in Table~\ref{tab:simulation_summarize}, will control the Type-I error under $10\%$.

We generate $400$ sequences to perform the detection procedures in each example. Meanwhile, all methods share reference data containing $400$ sequences from $f_0$ with lengths $15000$. The detection procedures allow us to learn the statistical behavior of the pre-change process from the reference data. For all NN-based methods, namely NN-CUSUM, ONNC, and ONNR, the model will train on a burn-in sequence consisting of pre-change data with a length of $5000$ to stabilize the neural networks before starting detection.

\vspace{.1in}
\noindent{\it Ten simulated examples.} Table~\ref{tab:simulated-distributions} briefly presents the ten examples in simulations. More details can be found in Appendix~\ref{app:additional_simulation}. We design examples to be challenging regarding small (or even zero) changes in mean and covariance. The data dimensionality is 100 in all simulated experiments.

\vspace{.1in}
\noindent{\it Baselines for comparisons.} The baselines include (1) Online Neural Network Classification (ONNC) and (2) Online Neural Network Regression (ONNR), (3) Window-limited Generalized Likelihood Ratio (WL-GLR), (4) Window-limited CUSUM (WL-CUSUM), (5) Multivariate Exponentially Weighted Moving Average (MEWMA), (6) Hotelling-CUSUM (H-CUSUM), and (7) Exact CUSUM. Exact CUSUM is included to provide a lower bound on EDDs for all the other baselines since it has full knowledge of both the pre- and post-change data distributions. The implementation details on the baselines can be found in Appendix~\ref{subsec:additional_baseline}. 

\begin{figure}[t!]
\centering 
\begin{subfigure}[h]{0.19\linewidth}
\includegraphics[width=\linewidth]{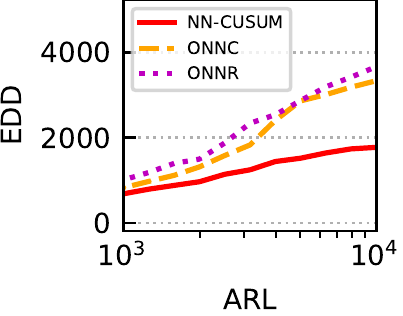}
\caption{Gaussian (mean)}
\end{subfigure}
\begin{subfigure}[h]{0.19\linewidth}
\includegraphics[width=\linewidth]{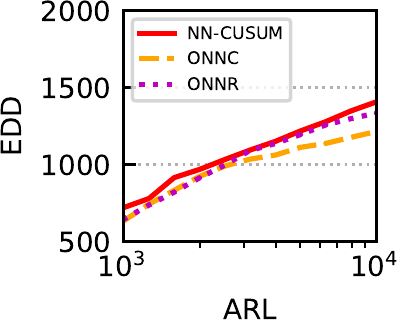}
\caption{Gaussian (cov.)}
\end{subfigure}
\begin{subfigure}[h]{0.19\linewidth}
\includegraphics[width=\linewidth]{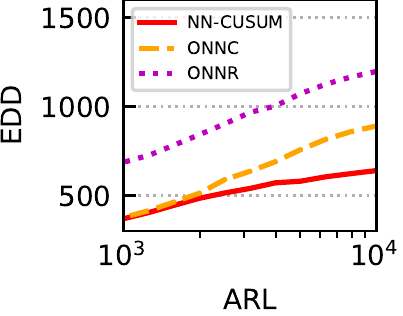}
\caption{Log Gaussian }
\end{subfigure}
\begin{subfigure}[h]{0.19\linewidth}
\includegraphics[width=\linewidth]{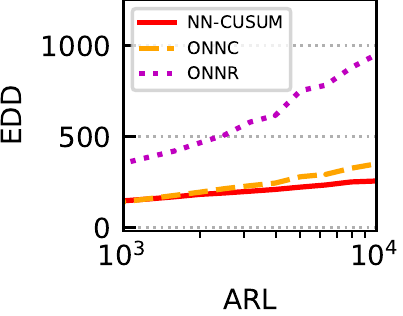}
\caption{GMM}
\end{subfigure}
\begin{subfigure}[h]{0.19\linewidth}
\includegraphics[width=\linewidth]{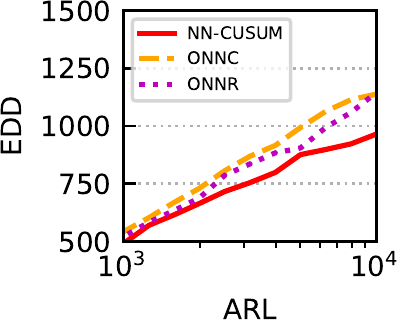}
\caption{Chi-square}
\end{subfigure}

\vspace{+10pt}
\begin{subfigure}[h]{0.19\linewidth}
\includegraphics[width=\linewidth]{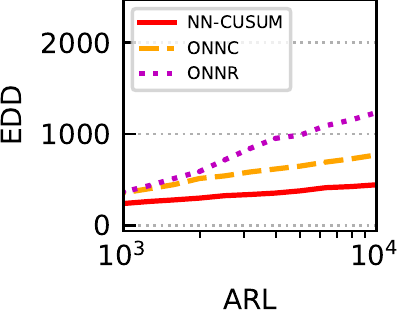}
\caption{Pareto}
\end{subfigure}
\begin{subfigure}[h]{0.19\linewidth}
\includegraphics[width=\linewidth]{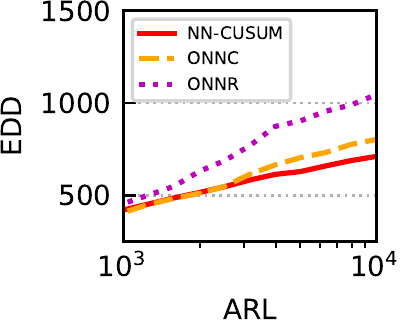}
\caption{Exponential}
\end{subfigure}
\begin{subfigure}[h]{0.19\linewidth}
\includegraphics[width=\linewidth]{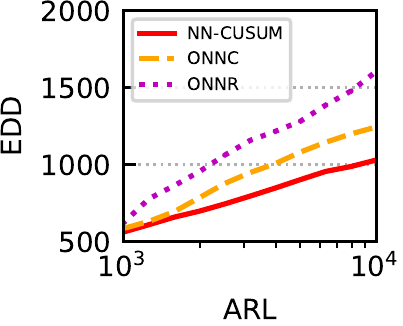}
\caption{Gamma}
\end{subfigure}
\begin{subfigure}[h]{0.19\linewidth}
\includegraphics[width=\linewidth]{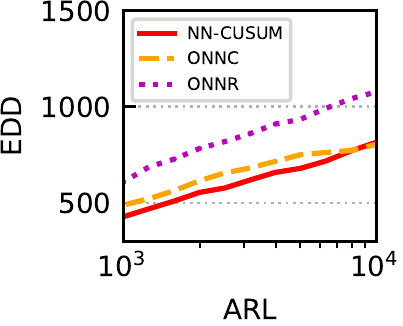}
\caption{Weibull}
\end{subfigure}
\begin{subfigure}[h]{0.19\linewidth}
\includegraphics[width=\linewidth]{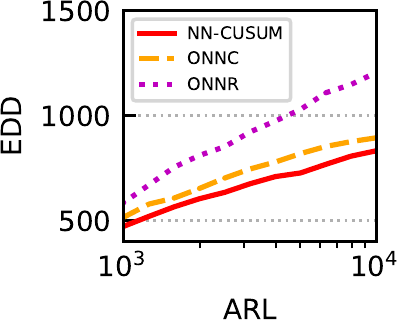}
\caption{Gompertz}
\end{subfigure}

\caption{ARL-EDD plots for NN-CUSUM, ONNC, and ONNR over ten examples are summarized in Table~\ref{tab:simulated-distributions}. The solid, dashed, and dotted curves are from NN-CUSUM, ONNC, and ONNR.}
\label{fig:compare-NNbased}
\label{fig:simulations}
\end{figure}
\begin{table}[t!]
    \caption{For $\ARL=5000$, EDDs on various methods and ten examples are reported with data in $\R^{100}$. The smallest EDD among the methods except for Exact CUSUM is bold in each setting. EDDs of Exact CUSUM are lower bounds for all methods. The standard deviations are in parentheses. The terms ``mean" and ``cov." indicate a mean and covariance shift, respectively. }
    \label{tab:simulation_summarize}
    \centering
    \begin{scriptsize}
\begin{tabular}{c ccccc}
\toprule
  
   & Gaussian (mean) & Gaussian (cov.) & Log Gaussian & GMM & 
     Chi-square \\
\midrule
NN-CUSUM 
 & {\bf 1520.31} (54.474)
 & 1218.15 (28.380)
 & {\bf 580.33} (12.791)
 & {\bf 222.75} (3.240)
 & {\bf 876.35} (20.864)
\\[1pt]
ONNC 
 & 2863.52 (85.842) 
 & 1111.88 (23.744) 
 & 757.78 (19.819)
 & 279.25 (6.650)
 & 993.52 (28.587)
\\[1pt]
ONNR 
 & 2869.97 (87.977) 
 & 1193.97 (23.848) 
 & 1071.28 (25.126)
 & 752.15 (16.905)
 & 903.10 (25.128)
\\[1pt]
WL-GLR 
& 3185.12 (89.588)
& 2305.72 (162.139)
& 1478.19 (65.042)
& 4962.48 (13.182)
& 3140.56 (85.305)
\\[1pt]
WL-CUSUM 
& 2789.28 (92.337)
& {\bf 921.05} (53.979)
& 1211.47 (46.487)
& 1160.42 (36.823)
& 2060.83 (69.974)
\\[1pt]
MEWMA 
& 2946.62 (88.818)
& 2317.50 (164.660)
& 1423.95 (62.799)
& 4177.16 (72.138)
& 3337.86 (87.842)
\\[1pt]
H-CUSUM 
& 2116.66 (73.193)
& 2054.87 (139.251)
& 919.05 (41.189)
& 5000.00 (0.000)
& 3825.60 (77.445)
\\[3pt]
Exact CUSUM  
& 358.93 (11.700)
& 14.21 (0.876)
& 1.02 (0.073)
& 1.00 (0.000)
& 58.52 (1.605)
\\
\bottomrule
\end{tabular}
\end{scriptsize}

\vspace{+10pt}

\begin{scriptsize}
\begin{tabular}{c c@{\hskip 15pt}c@{\hskip 15pt}c@{\hskip 15pt}c@{\hskip 15pt}c}
\toprule
  
   & Pareto & Exponential & Gamma & Weibull & 
     Gompertz \\
\midrule
NN-CUSUM  
& {\bf 377.05} (5.970)
& {\bf 628.60} (11.055)
& {\bf 902.42} (20.25)
& {\bf 679.68} (13.228)
& {\bf 726.25} (14.27)
\\[1pt]
ONNC 
& 647.60 (13.165)
& 703.75 (14.218)
& 1081.50 (23.477)
& 749.70 (13.187)
& 818.48 (16.053)
\\[1pt]
ONNR 
& 986.25 (21.686)
& 903.05 (18.854)
& 1281.68 (31.480)
& 935.38 (16.706)
& 1029.03 (19.331)
\\[1pt]
WL-GLR 
& 4877.48 (31.336)
& 4968.20 (15.052)
& 4988.06 (9.359)
& 5000.00 (0.000)
& 5000.00 (0.000)
\\[1pt]
WL-CUSUM 
& 423.36 (4.500)
& 5000.00 (0.000)
& 5000.00 (0.000)
& 5000.00 (0.000)
& 5000.00 (0.000)
\\[1pt]
MEWMA 
& 4901.52 (28.060)
& 4996.67 (3.323)
& 5000.00 (0.000)
& 5000.00 (0.000)
& 5000.00 (0.000)
\\[1pt]
H-CUSUM 
& 4908.90 (27.127)
& 5000.00 (0.000)
& 5000.00 (0.000)
& 5000.00 (0.000)
& 5000.00 (0.000)
\\[3pt]
Exact CUSUM  
& 2.79 (0.079)
& 1.00 (0.000)
& 1.00 (0.000)
& 1.00 (0.000)
& 1.00 (0.000)
\\
\bottomrule
\end{tabular}
\end{scriptsize}
\end{table}

\vspace{.1in}
\noindent{\it Comparisons with NN-based methods.} Figure~\ref{fig:compare-NNbased} compares the three NN-based methods, and NN-CUSUM detects quicker except for Gaussian covariance shift in all experiments. In this exception, the performance of NN-CUSUM is close to ONNC and ONNR. NN-CUSUM can outperform the other two NN-based methods because the change-point detection tasks are designed to be difficult. To be concrete, we set the change to be small by using small magnitudes, employing sparsity, or fixing the expectation. Though ONNC and ONNR are supposed to detect any distributional shift, they are Shewhart chart types (see, e.g., \cite{xie2021sequential}) that are less sensitive to small changes. Thus, they detect slower than NN-CUSUM.

\vspace{.1in}
\noindent{\it Comparisons with classic methods.}
Table~\ref{tab:simulation_summarize} summarizes the EDDs of all methods for a fixed ARL at $5000$. Overall, NN-CUSUM outperforms other baselines with smaller detection delays. As for classic methods, when their own assumptions are not violated, they can detect the change, but the detection delay relies on how significant the change is over the noise. Like in Gaussian sparse mean shift and Gaussian sparse covariance shift, all methods can more or less detect the change. However, given the sparsity and small magnitude of the shifts, the detection can be slow. Indeed, the classic methods considered in our experiments rely on the first and second-order moment statistics to capture the distribution shift. Thus, the classic methods can sometimes work even when their assumptions are violated, but the changes in expectations or variances are evident. For example, WL-GLR and WL-CUSUM are acute to mean shifts, and H-CUSUM is sensitive to the increase in variances. In the simulations, however, the expectation and variance changes are small (or even no change), leading to frequent failures of classic methods.


\subsection{Real data experiments}

\begin{figure}[b!]
\centering
\begin{subfigure}[h]{0.4\linewidth}
\includegraphics[width=\linewidth]{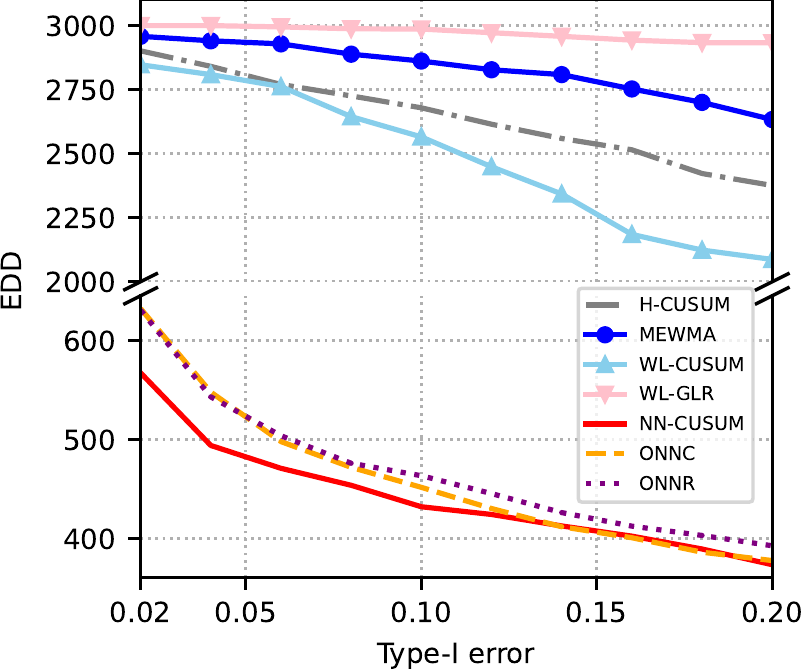}
\caption{Higgs boson}
\label{fig:higgs_boson_EDD}
\end{subfigure}
\hspace{+30pt}
\begin{subfigure}[h]{0.4\linewidth}
\includegraphics[width=\linewidth]{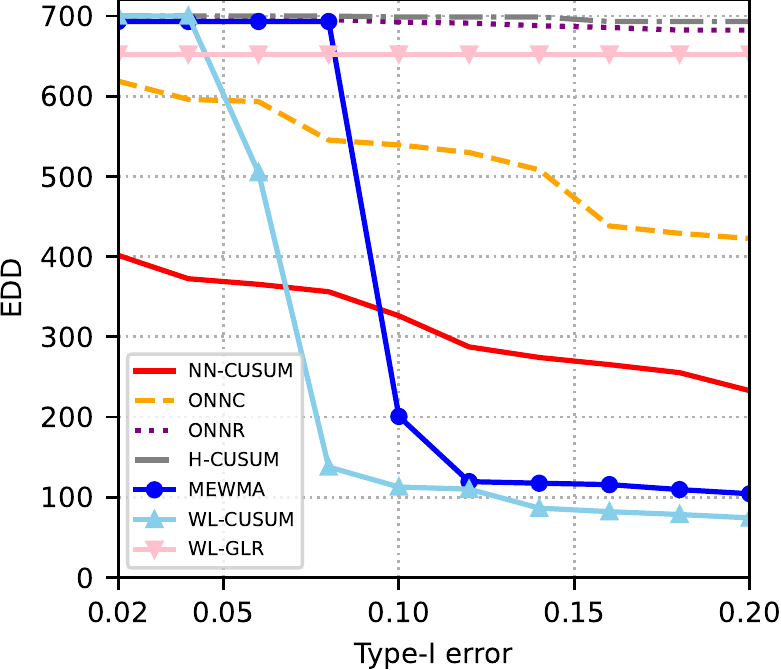}
\caption{MiniBooNE}
\label{fig:miniboone_edd}
\end{subfigure}

\caption{EDD versus Type-I errors in the range of $\{0.02, 0.04 , \ldots , 0.20 \}$ on (a) Higgs boson dataset and (b) MiniBooNE dataset.}
\label{fig:real_data_EDD}
\end{figure}

We also evaluate the performance of NN-CUSUM and baselines for change-point detection using two real-world datasets: Higgs boson dataset and MiniBooNE dataset.

\vspace{.1in}
\noindent{\it Dataset.} Collisions at high-energy particle colliders are useful for discovering rare particles such as Higgs boson. Machine learning approaches often detect such rare particles, which requires solving challenging signal-versus-background classification problems~\cite{baldi2014searching}. In our experiment, we convert the Higgs boson detection task into a change-point detection task using the Higgs boson dataset to evaluate NN-CUSUM and baselines. The Higgs boson dataset contains data samples obtained from two types of signals: Higgs boson signal and background signal that does not produce Higgs bosons. The dataset contains about ten million samples produced from Monte Carlo simulations. Each sample comprises 28 numeric features, where 21 features are kinematic properties measured by the particle detectors, and seven features are functions of the first 21 features, which are high-level features derived from the 21 kinematic features by physicists. In our experiment, we use all 28 features.

Detecting neutrinos is difficult due to their low mass and lack of electrical charge~\cite{2020NeutrinosCS}. Mini Booster Neutrino Experiment (MiniBooNE) is designed to search for neutrino oscillations, which provides the MiniBooNE dataset. Similar to the Higgs boson dataset, we establish a change-point detection task using the MiniBooNE dataset. The dataset contains about 130 thousand data samples obtained from background signals or electron neutrinos. Each sample has 50 numeric features, all used in our experiment.

For illustration, we refer to the non-background signal in each dataset as the target signal throughout this section.
Both datasets are available at the University of California Irvine (UCI) machine learning repository~\cite{2024UCI}.

\vspace{.1in}
\noindent{\it Baselines.} We use the same baselines as in simulations: WL-GLR, WL-CUSUM, H-CUSUM, MEWMA, ONNC, and ONNR. Exact CUSUM is excluded since we do not know the true distribution of the real data.

\vspace{.1in}
\noindent{\it Setup.} We construct reference and online sequences for each dataset where only online sequences have a change point. We create $500$ reference and online sequences using the Higgs boson dataset: one reference sequence contains $3500$ data samples from the background signal; one online sequence contains $500$ data samples from the target signal and $3000$ data samples from the background signal so that the change point occurs at time index $500$ in online sequence. We also construct $100$ reference and online sequences using the MiniBooNE dataset similarly: one reference sequence contains $1200$ data samples from background signal; one online sequence contains $500$ data samples from background signal and $700$ data samples from target signal so that the change point occurs at time index $500$.

We use the same neural network architecture for NN-CUSUM, ONNC, and ONNR: a fully connected neural network with one hidden layer of 1024 width and ReLU activation. The window size is set to 100 for all methods that require a window. Training batch size and stride of the sliding window are set to 10 for NN-CUSUM, ONNC, and ONNR. We set a burn-in sequence of length 500 for NN-CUSUM, ONNC, and ONNR. Adam is used for training the neural network of NN-CUSUM, ONNC, and ONNR. The learning rate is set to $10^{-3}$ without scheduling.

\vspace{.1in}
\noindent{\it Type-I error.} ARL may not be a good metric for false-alarm control in accurate data since we are interested in the capability of detecting a change quickly within a time frame. Thus, we use a Type-I error as an alternative to the ARL, which is defined as $\bbP_0[\tau\le k]$ representing the probability that a detection procedure raises an alarm before the change point $k$ under $f_0$. With an arbitrarily fixed $b$, the detection procedure stops before $k+1$ is equivalent with the event $\{\max_{t\le k}S_t > b\}$. In other words, Type-I errors can be approximated by the empirical frequency of $\{\max_{t\le k}S_t > b\}$ over sequences.


\begin{table}[t!]
\caption{Detection failure rate with $0.02$, $0.10$, and $0.20$ Type-I error on Higgs boson and MiniBooNE dataset. The lowest detection failure rate for each Type-I error is in bold.}
\scriptsize
\label{table:failure_rate}
 \vspace{-10pt}
\begin{center}
\begin{scriptsize}

\begin{tabular}{ccccccc}
\toprule
 & \multicolumn{3}{c}{Higgs boson}  & \multicolumn{3}{c}{MiniBooNE}\\
 \midrule
 & \multicolumn{3}{c}{Type-I Error}  & \multicolumn{3}{c}{Type-I Error}\\
 & 0.02 & 0.10 & 0.20 & 0.02 & 0.10 & 0.20\\
\midrule
NN-CUSUM & \bf{0.00}  & \bf{0.00} & \bf{0.00} & \bf{0.30} & \bf{0.20} & \bf{0.12} \\ 
ONNC & \bf{0.00}  & \bf{0.00} & \bf{0.00}  &  0.94  & 0.67 & 0.41 \\
ONNR &  \bf{0.00} & \bf{0.00} & \bf{0.00} & 0.88 & 0.42 & 0.21\\
WL-GLR & 1.00 & 0.99 & 0.97  &  1.00  & 1.00 & 0.96\\
WL-CUSUM & 0.89 & 0.71 & 0.46 & 0.89  & 0.77 & 0.46 \\
MEWMA & 0.97 & 0.91 & 0.77 & 0.97  & 0.91 & 0.77  \\
H-CUSUM & 0.92  & 0.78 & 0.62 & 0.91 & 0.76 & 0.61 \\

\bottomrule
\end{tabular}
\end{scriptsize}
\end{center}
\end{table}

\vspace{.1in}
\noindent{\it Results.} 
Figure~\ref{fig:higgs_boson_EDD} shows EDD versus multiple Type-I errors for NN-CUSUM and baselines on the Higgs boson dataset. Type-I error takes value in $\{0.02,0.04, \ldots, 0.18,0.20 \}$. We observe that NN-CUSUM consistently outperforms all other baselines with Type-I errors in $\{0.02,\ldots,0.12\}$. We also evaluate how often a detection procedure fails to detect the change point by calculating the detection failure rate. The detection failure rate is defined by the frequency of the event $\{\max_{t> k}S_t \le b\}$ over sequences. Table~\ref{table:failure_rate} shows the detection failure rate of NN-CUSUM and baselines with respect to Type-I errors in $\{0.02, 0.10, 0.20\}$ on the Higgs boson dataset. NN-CUSUM, ONNC, and ONNR achieve $0$ detection failure rate over all Type-I errors.

Figure~\ref{fig:miniboone_edd} shows EDD versus Type-I errors in the range of $\{0.02,0.04, \ldots, 0.18,0.20 \}$. While NN-CUSUM outperforms other baselines with Type-I errors in $\{0.02,0.04,0.06\}$, WL-CUSUM and MEWMA outperform NN-CUSUM in a higher Type-I error. NN-CUSUM, however, consistently shows the lowest detection failure rate of all other baselines, as shown in Table~\ref{table:failure_rate}.

\section{Conclusion}
\label{sec:conclusion}
We present a new neural network-based procedure for online change-point detection. We use a CUSUM recursion with increments obtained from a trained neural network binary classifier applied on test batches. The motivation is that the training of a logistic loss will converge (in population) to a log-likelihood ratio between two samples, which thus naturally motivated us to construct CUSUM statistics using the sequentially learned neural network to test samples. We further generalize it to other training losses, such as the MMD loss, which enables the neural tangent kernel (NTK) analysis. We develop a novel online procedure by training and testing data-splitting to allow us to explain the theoretical performance of the proposed NN-CUSUM procedure. 
Using extensive numerical experiments, we demonstrate the competitive performance of the new procedure compared with the classical statistical method and other neural network-based procedures to show its good performance on simulated and real data experiments.


\section*{Acknowledgement}

This work is supported by  NSF DMS-2134037.
TG, JL, and YX are also
 partially supported by NSF CAREER CCF-1650913, CMMI-2015787, DMS-1938106, and DMS-1830210.
 XC is also partially supported by 
 NSF DMS-2031849 and
 the Simons Foundation. 
\bibliographystyle{plain}
\bibliography{refs}

\appendices



\section{Proofs}

\begin{proof}[Proof of Lemma \ref{lemma:conc-test}]
Let $g=g_{\hat \theta}$ be fixed, $ \sup_{x\in\calX} |g(x)| \le C$.
By definition, with $m' = (1-\alpha) w $, 
\[
 \eta_t = \frac{1}{m'} \sum_{i=1}^{m'} ( g( x_i) - g ( \tilde{x}_i) ),
\]
where $x_i \sim f_1$, i.i.d,. $\tilde{x}_i \sim f_0$, i.i.d., and are independent.
The the random variable $( g( x_i) - g ( \tilde{x}_i) )$ is i.i.d. and is bounded by $2C$. 
The lemma directly follows Classical Hoeffding’s inequality.
\end{proof}

\begin{proof}[Proof of Lemma \ref{lemma:NTK-etat-bound}]
The proof follows a similar argument in \cite[Lemma B.1]{cheng2021neural}. 
By definition \eqref{eq:def-eta-t-exactNTK},\[
\bar{\eta}: =
\E_1 [ \eta_t^{\rm NTK} | D_t^{\rm tr}]
= \int (\phi_1(x') - \phi_0(x')) (\hat{f}_{1,t}^{\rm tr} (x')- \hat{f}_{0,t}^{\rm tr} (x')) dx',
\]
where
\[
\phi_i(x') := \int K_0(x, x') f_i(x)dx, \quad i=0,1.
\]
Thus $\bar{\eta}$ can be written as the independent sum over the training samples as
\[
\bar{\eta} = \frac{1}{m} \sum_{i=1}^m (\phi(x_i) - \phi(\tilde{x}_i)),
\quad \phi:= \phi_1 - \phi_0,
\]
where  $x_i \sim f_1$, i.i.d,. $\tilde{x}_i \sim f_0$, i.i.d. are from the training split and are independent, $m = \alpha w$.
Meanwhile, by the uniform boundedness of $K_0(x,x') \in [0,B]$,
\[
\phi_i(x) \in [0,B], \quad \forall x \in \calX, \quad i=0,1.
\]
As a result,
\[
| \phi(x) | = |\phi_1(x) - \phi_0(x)| \le B, \quad \forall x\in \calX.
\]
Then, the classical Hoeffding inequality gives that for any $\lambda_2>0$ 
\[
 \mathbb{P}_1 \left[ \bar{\eta} - \E_1 \bar{\eta} <- 
  \frac{\lambda_2 B}{ \sqrt{\alpha w}}  \right] 
\le e^{ - \lambda_2^2/2},
\]    
where the probability $ \mathbb{P}_1$ is over the randomness of the training data $D_t^{\rm tr}$ post change.
Recall that $ \E_1 \bar{\eta}  =  {\rm MMD}^2( f_1, f_0)$ by definition, 
the lemma follows.
\end{proof}

\begin{proof}[Proof of Lemma \ref{CUSUM_finite}]
Let $r > b/\delta$. Note that for any length-$r$ consecutive observations, since for some $\delta \in (0, 1)$, $\mathbb P\{X_1\geq \delta\}\geq \delta$, we have $\mathbb P\{X_{n} + \cdots + X_{n+r-1} > b\} \geq \delta^r$ for arbitrary $n\geq 1$. Now break the sequence into length-$r$ segments, and define a ``geometric'' random variable 
\[
M = \mbox{first value $m'r ~(m'\geq 1)$  such that } \sum_{n=(m'-1)r + 1}^{m'r} X_n > b.
\]
Then it can be shown that for $m' =1, 2, \ldots$
\[
\mathbb P\{N > m'r\} \leq \mathbb P \{M > m'r\},
\]
since 
\[\{M \leq m'r\} \subseteq \{\sum_{n=(m-1)r+1}^{mr} X_n > b, m \leq m'\} \subseteq \{\max_{1\leq t\leq m'r }\max_{1<k<t} \sum_{n=k+1}^{t} X_n > b\}=\{N \leq m' r\}.\]
On the other hand, $\mathbb P \{M > m'r\} < (1-\delta^r)^{m'}$. Finally, we have
\begin{equation}
    \begin{split}
        \mathbb E[N] &= \sum_{n=1}^\infty \mathbb P\{N > n\}\\
        & = \sum_{m'=1}^\infty \sum_{n=(m'-1)r+1}^{m'r} \mathbb P\{N > n\}\\
        & \leq \sum_{n=1}^r \mathbb P\{N > n\}+\sum_{m'=1}^\infty \mathbb P\{N > m'r\}\\
        & \leq r + \sum_{m'=1}^\infty (1-\delta^r)^{m'} = r+(1-\delta^r)/\delta^r <\infty.
    \end{split}
\end{equation}
\end{proof}

\begin{proof}[Proof of Theorem \ref{main_thm}]

Consider the stopping time $\tau$ for NN-CUSUM defined in \eqref{eq:NNCUSUM}. Due to absolute integrable assumption, $\eta_t$ satisfies (i) in Lemma \ref{Walds}. And since $\tau$ is a stopping time, so it also satisfies (ii). So we only need to check (v) in Lemma \ref{Walds} (which, together with absolute integrability of $\eta_t$, can lead to (iii)). 

Now, to check check (v) in Lemma \ref{Walds}, we will use Lemma \ref{CUSUM_finite} for NN-CUSUM; we just need that after the change, under $f_1$, the i.i.d. increments to have a non-negligible chance to be positive: $\mathbb P_1(\eta_t - D>\delta) > \delta$ for some small constant $\delta >0$. As a result of Lemma \ref{lemma:conc-test}, and Lemma \ref{lemma:NTK-etat-bound}, when $w$ is sufficiently large and on the order of $O(1/({\rm MMD}^2(f_1, f_0)^2)$, we have the increment $\eta_t$ from training MMD loss satisfies \eqref{increment_high_prob} with high probability. Hence, the NN-CUSUM stopping time can satisfy $\mathbb E_1[\tau]< \infty$. Thus, we can invoke Wald's identity Lemma \ref{Walds} for a stopped process induced by $\tau$.  Since $\min_{n=0}^t \tilde S_n\leq 0$, so $S_t \geq \tilde{S}_t$. So we have at the stopping moment, $S_\tau \geq \tilde S_\tau$. Since $\tau$ is a stopping time that can satisfy $\mathbb E_1[\tau]< \infty$ and, hence, we can apply Walds's identity Lemma \ref{Walds} to obtain the following
\begin{equation}\label{EDD_bound}
    \mathbb E_1 [\tau] = \frac{\mathbb E_1[\tilde S_\tau]}{\mathbb E_1[\eta_t]-D} 
    \leq 
    \frac{\mathbb E_1[S_\tau]}{\mathbb E_1[\eta_t]-D} = \frac{b + \mathbb E_1[S_\tau - b]}{\mathbb E_1[\eta_t] - D},
\end{equation}
where the first equality is due to Wald's identity Lemma \ref{Walds} for $\tilde S_t$ with stopping time defined by $\tau$. From the last inequality in \eqref{EDD_bound}, we obtain \eqref{EDD_approx} by using \eqref{positive_exp}.
\end{proof}
Note that, although in the proof we have shown $\mathbb E_1[\tau]$, even before change occurs and under $f_0$, as long as we also have $\mathbb P_0(\eta_t - D>\delta) > \delta$, then we also have $\mathbb E_0[\tau]< \infty$, i.e., the expected stopping time will be finite, meaning that there will be false alarms. 

\begin{proof}[Proof of Lemma \ref{generalized_walds}]
 We observe that for $t > m$, we have
\begin{equation}\label{break-down}
\begin{split}
\mathbb E [\sum_{n=m+1}^{N} X_n] 
& = \underbrace{\mathbb E \left [ \sum_{n=m+1}^{m+N} X_n\right]}_{(a)}
-  \underbrace{\mathbb E \left[\sum_{n=N+1}^{m+N}X_n\right]}_{(b)}
\end{split}
\end{equation}
We now analyze the two terms on the right-hand-side above separately:
\begin{equation}
\begin{split}
(a) &= \mathbb E\left[\sum_{n=m+1}^{m+N} X_n \right] \\
&=\mathbb E\left[\sum_{n=m+1}^{\infty} X_n \mathbb I\{N \geq n-m\} \right]\\
&=\mathbb E \left[\sum_{n=m+1}^{\infty} \mathbb E[X_n|\mathcal F_{n-m-1}] \mathbb I\{N \geq n-m\}\right]\\
&=\mathbb E \left[\sum_{n=m+1}^{\infty} \mathbb E[X_n] \mathbb I\{N \geq n-m\}\right]
\end{split}
\end{equation}
the second equation is because $X_n$ is $\mathcal F_{n-m-1}$-measurable, since it is $m$-dependent. Thus, we have
\[
(a) = \mathbb E[N] \mathbb E[X_1]
\]
For the second term, we have
\begin{equation}
\begin{split}
(b) &= \mathbb E \left[\sum_{n=N+1}^{N+m} X_n\right]\\
&= \mathbb E \left[\sum_{n=m+1}^{\infty} X_n \mathbb I\{n > N\} \mathbb I\{N \geq n-m\}\right] \\
& =\mathbb E\left[\sum_{n=m+1}^{\infty} \mathbb E[X_n|\mathcal F_{n-1}] \mathbb I\{n > N\} \mathbb I\{N \geq n-m\}\right]\\
& =\mathbb E\left[\mathbb E[X_{m}|\mathcal F_{m-1}]\sum_{n=m+1}^{\infty} \mathbb I\{n > N\} \mathbb I\{N \geq n-m\}\right]
\end{split}
\end{equation}
where the third equality is due to the fact that $\mathbb I\{n>N\}$ is $\mathcal F_{n-1}$-measurable, and $\mathbb I\{N \geq n-m\}$ is $\mathcal F_{n-1-m}$-measurable and thus also $\mathcal F_{n-1}$-measurable, since $\mathcal F_{n-m-1}\subseteq \mathcal F_{n-1}$.
Then, we have 
\[(b) = m \mathbb E[X_1].\]

Now, consider the left-hand-side of \eqref{break-down}
$\mathbb E [\sum_{n=m+1}^{N} X_n] =\mathbb E [\sum_{n=m+1}^{N} X_n-b] +b.$
At the stopping time $N$, the overshoot $\mathbb E[\sum_{n=m+1}^{N} X_n- b]$ can be estimated as typically on the order of the expected increment $\mathbb E[X_1]$ (see,  e.g., \cite{xie2022window} for an analysis of the expected increment). The rest of the lemma follows from single arguments.
\end{proof}

\section{Additional simulation results}\label{app:additional_simulation}

\subsection{Details on baselines}\label{subsec:additional_baseline}

In this section, we introduce the technical details of the compared baselines. As a general scheme for change-point detection, the practitioners need to calculate the statistics at each time and compare the statistics with a pre-specified threshold. When the sequential statistics exceed a threshold at some time $t$, an alarm will be raised to claim a change happens, and the stopping time $\tau$ will be set to the value $t$. 

\vspace{.1in}
\noindent{\it Online Neural Network Classification (ONNC).} ONNC is an online change-point detection method using neural networks proposed by Hushchyn et al.~\cite{hushchyn2020online}. The authors assume no reference data in their context and use historical data for the negative class data, potentially affecting their detection efficiency due to a non-zero chance that the historical data is (partially) positive class. To make fair comparisons, we modify the method by training their model with negative class data from reference data to fit our context. The neural network of ONNC is trained with the logistic loss function to learn the labels directly.
The test statistic at time $t$ is calculated as
\[
\eta_t^{\ONNC} = -\frac{1}{(1-\alpha)w} \left[ \sum_{x \in X_t^{{\rm te}}} \log\left(\frac{1-g_{\hat\theta_t}^\ONNC(x)}{g_{\hat\theta_t}^\ONNC(x)}\right) -
\sum_{x \in \widetilde{X}_t^{\rm te}} \log\left(\frac{g_{\hat\theta_t}^\ONNC(x)}{1-g_{\hat\theta_t}^\ONNC(x)}\right) \right],
\]
where $\alpha$ is split ratio, $w$ is window size and $g_{\theta_t}^\ONNC(x)$ is the test function of ONNC at time $t$. ONNC method raises the alarm under the stopping time $\tau_{\rm ONNC} = \inf\{t: \eta_t^\ONNC > b\}$.

\vspace{.1in}
\noindent{\it Online Neural Network Regression (ONNR).} ONNR is also a change-point detection method using neural networks proposed by Hushchyn et al.~\cite{hushchyn2020online}. The difference from ONNC is that ONNR learns the density ratio between pre- and post-change data rather than learning the binary labels. The finite-sample loss function in ONNR on the training set is defined as
\[
l^{\rm ONNR}\left(\theta_{t};\widetilde X_t^{\rm{tr}}, X_t^{\rm{tr}}\right) = \frac{1-a}{2m} \sum_{x \in \widetilde X_t^{\rm{tr}}} \left(g^\ONNR_{\theta_{t}}(x)\right)^2 + \frac{a}{2m} \sum_{x \in X_t^{\rm{tr}}}  \left(g^\ONNR_{\theta_{t}}(x)\right)^2 - \frac{1}{m} \sum_{x \in X_t^{\rm{tr}}}  g^\ONNR_{\theta_{t}}(x),
\]
where $m$ is training size, $a \in (0,1)$ is a hyperparameter and $g_{\theta_t}^\ONNR(x)$ is test function for ONNR. 
Due to the asymmetry with respect to $\widetilde X_t^{\rm{tr}}$ and $X_t^{\rm{tr}}$ in the loss function, ONNR uses two separate neural networks to train over two loss functions, $l^{\rm ONNR}\left(\theta_{1,t};\widetilde X_t^{\rm{tr}},  X_t^{\rm{tr}}\right)$ and $l^{\rm ONNR}\left(\theta_{2,t}; X_t^{\rm{tr}}, \widetilde X_t^{\rm{tr}}\right)$, respectively. The test statistic at time $t$ is calculated as 
\[
\eta_t^\ONNR = \frac{1}{(1-\alpha)w} \sum_{x \in X_t^{\rm{te}}} \left( g_{\hat\theta_{1,t}}^\ONNR(x) - 1 \right) + \frac{1}{(1-\alpha)w} \sum_{x \in \widetilde X_t^{\rm{te}}} \left( g_{\hat \theta_{2,t}}^\ONNR(x) - 1 \right),
\]
where $\alpha$ is split ratio and $w$ is window size.
ONNR method raises the alarm under the stopping time $\tau_{\rm ONNR} = \inf\{t: \eta_t^\ONNR > b\}$.

\vspace{.1in}
\noindent{\it Window-Limited Generalized Likelihood Ratio (WL-GLR).} 
For WL-GLR, we assume $f_0$ is known and $f_1(\cdot,\theta)$ is partially known with a parametric form and an unknown parameter $\theta\in\Theta$. With a window size $w$, the sequential statistics of WL-GLR are given by
\[
S_t^{\rm WLG}=\max _{(t-w)^+ \leq i < t} \sup _{\theta \in \Theta} \sum_{j=i}^t \ell(x_j,\theta), 
\]
where $\ell(x_t,\theta) = \log \left(f_1(x_t,\theta)/f_0(x_t)\right)$. In general, no recursion can be yielded in the GLR algorithm. Therefore, the computation cost of the GLR algorithm is huge. However, the computation is tractable when we assume the problem is Gaussian mean shift. Then, the statistics will degenerate to
\[
S_t^{\rm WLG}=\max _{(t-w)^+ \leq i < t} \frac{\left[\sum_{j=i+1}^t(x_j-\hat{\mu}_p)\right]^\top\widehat\Sigma_p^{-1}\left[\sum_{j=i+1}^t(x_j-\hat{\mu}_p)\right]}{t-i},
\]
where $\hat{\mu}_p$ and $\widehat\Sigma_p$ are estimated mean and covariance from reference data. We use the degenerated WL-GLR throughout the experiments. The stopping time is defined as $\tau_{\rm WLG} = \inf\{t: S_t^{\rm WLG} > b\}$.

\vspace{.1in}
\noindent{\it Window-Limited CUSUM (WL-CUSUM).} 
WL-CUSUM is proposed in \cite{xie2022window}. When the pre-change density $f_0$ is fully known and the post-change density $f_1(\cdot,\theta)$ has a parametric form but an unknown parameter $\theta\in\Theta$, the log-likelihood ratio is adaptively defined as
\[
\ell(x_t,\hat\theta_{t-1}) = \log \frac{f_1(x_t,\hat\theta_{t-1})}{f_0(x_t)},
\]
where $\hat\theta_{t-1}\in\Theta$ is an estimation of $\theta$. If a window size $w$ is fixed, the estimation $\hat\theta_t$ relies on the data within the window $\{x_t,\ldots,x_{t-w+1}\}$. The option on the estimator is not rigid. The most common option is the Maximum Likelihood Estimator (MLE) taking the form:
\[
\hat{\theta}_t=\argmax _{\theta \in \Theta} \sum_{i=0}^{w-1} \log f_1\left(x_{t-i}, \theta\right).
\]
Throughout the experiments, we assume $f_0$ and $f_1$ are parameterized by Gaussian distribution and $\hat\theta_t$ is estimated by MLE. With the initial value $S_0^{\rm WLC}=0$, the recursion takes the form
\[
S_t^{\rm WLC} = \max\left\{S_{t-1}^{\rm WLC} + \ell(x_t,\hat\theta_{t-1}),0\right\}.
\]
The stopping time is determined by $\tau_{\rm WLC} = \inf\{t: S_t^{\rm WLC} > b\}.$

\vspace{.1in}
\noindent{\it Multivariate Exponentially Weighted Moving Average (MEWMA).} 
MEWMA calculates a statistic that combines the differences between the current observation and the expected values based on historical data, weighted by exponentially decreasing weights. This weighting scheme assigns more importance to recent observations while still considering the entire history. As introduced in \cite{lowry1992multivariate}, MEWMA computes
$$
z_t=r x_t+(1-r)z_{t-1},\quad t\ge 1
$$
where $z_0 = 0$ and the decay rate belongs to the range $0<r\le 1$. The MEWMA chart raises the alarm under the stopping time
$$
\tau_{\rm M} = \inf\{t: z_t^\top \Sigma_{z_t}^{-1} z_t > b\}
$$
where the covariance matrix of pre-change $z_t$ is specified by $\Sigma_{z_t} = \left(r\left(1-(1-r)^{2 t}\right) /(2-r)\right) \Sigma_0$. 

\vspace{.1in}
\noindent{\it Hotelling CUSUM (H-CUSUM).} 
Instead of using the exact log-likelihood ratio in the recursion, H-CUSUM procedure uses the Hotelling T-square statistic in the recursion. It can be treated as a non-parametric detection statistic (only uses the first and the second order moments). It is good in detecting mean shifts but not very effective in detecting other types of changes (such as covariance shifts). Define 
\[
\begin{split}
g_0^{\rm H}(x_t) 
& = \frac{1}{2}(x_t- \hat{\mu}_p)^\top \left(\widehat{\Sigma}_p + \nu I_d\right)^{-1} (x_t- \hat{\mu}_p),\\
g^{\rm H}(x_t) 
& = g_0^{\rm H}(x_t) - \hat{d}_p
\end{split}
\]
where $\hat{\mu}_p$ and $\widehat{\Sigma}_p $ are estimated mean, and covariance matrix from the reference sequence assumed to be drawn from $f_0$;
$\hat{d}_p$ usually set to be  $\E_{x \sim f_0}  g^{\rm H}_0(x) + \epsilon$ for some small constant $\epsilon$ 
and is computed by sample average on a test split of the reference sequence in practice. 
The positive scalar $\nu$ is a regularizing parameter because the covariance matrix estimator $\widehat{\Sigma}_p$ may be singular, e.g., when the data dimension is high compared to the size of the reference sequence. 
Once $\hat{\mu}_p$, $\widehat{\Sigma}_p $ and $\hat{d}_p$ are pre-computed on the reference set, the recursive CUSUM is computed upon the arrival of stream samples of $x_t$ as 
\[
S^{\rm H}_t = \max\left\{ S^{\rm H}_{t-1} + g^{\rm H}(x_t), 0 \right\}.
\]
The associated stopping time procedure is $\tau_{H} = \inf\{t: S^{\rm H}_t > b\}$.

\subsection{Details in simulated examples.}

A brief summary of settings is given in Table~\ref{tab:simulated-distributions}. In the following, the concrete settings are  
\begin{enumerate}[label=(\arabic*)]
\item {\it Gaussian sparse mean shift:} We set the shift $\mu_q = (\delta, \delta/2, \delta/3, 0, \cdots, 0)^\top$ and the shift magnitude $\delta=0.1$; 
\item {\it Gaussian covariance shift:} $E$ is $d \times d$ all-ones matrix, and $D$ is a diagonal matrix, defined as $D = \sqrt{\rho} D_0$ with $D_0$ a sparse diagonal matrix in which 20 out of 100 of the diagonal entries are ones, and the remaining entries are zeros. The index set of non-zero diagonal entries is $\{1,6,\ldots,91,96\}$. We set the covariance shift magnitude $\rho=0.1$.
\item {\it Log Gaussian distribution with covariance shift:} We shift the covariance matrix from $I_d$ to $0.8I_d+0.2E$. 
\item {\it Gaussian Mixture Model (GMM) component shift:} For post-change distribution, we set $\mu_q = 0$ and $\Sigma_q = 0.8I_d+0.2E$ where $I_d$ is a $d$-by-$d$ identity matrix and $E$ is a $d$-by-$d$ all-one matrix.
\item {\it Non-central Chi-square distribution with non-centrality shift:} We set the degree of freedom $\kappa_c=0.5$, and change non-centrality parameters from $\nu_0 = 1$ to $\nu_1 = 0.6$ for sparse coordinates in $\{1,26,51,76\}$. 
\item {\it Pareto distribution with shape shift:} The density of Pareto distribution is
\[
f_{\text{Pareto}}(x ; x_m, b) = \frac{bx_m^b}{x^{b+1}},\quad x\ge x_m>0,\quad b>0, 
\]
where $x_m$ is the lower bound of variables, and $b$ is the shape parameter.
We set the lower bound $x_m=1$, and shift shape parameters from $b_0 = 2$ to $b_1 = 2.5$.
\item {\it Exponential distribution with scale and location shift:} The density of Exponential distribution is
\[
f_{\text{Exp}}(x ; \beta) = \frac{1}{\beta} e^{-x / \beta},\quad x \geq 0,
\]
where $\beta$ is the scale parameter.
We change scale parameters from $\beta_0 = 1$ to $\beta_1 = 0.8$ and set the location shift $\mu_q = \beta_0-\beta_1$.
\item {\it Gamma distribution with scale and location shift:} The density of Gamma distribution is
\[
f_{\text{Gamma}}(x ; \kappa, \beta)=\frac{x^{\kappa-1} e^{-x / \beta}}{\beta^\kappa \Gamma(\kappa)},\quad x, \kappa, \beta>0,
\]
where $\kappa$ is the shape parameter and $\beta$ is the scale parameter. 
We change scale parameters from $\beta_0 = 0.5$ to $\beta_1 = 0.4$, set shape parameter $\kappa=1.5$, and the location shift $\mu_q = (\beta_0-\beta_1)\kappa$.
\item {\it Weibull distribution with scale and location shift:} The density of Weibull distribution is
\[
f_{\text{Weibull}}(x ; \kappa,\beta) = \frac{\kappa}{\beta}\left(\frac{x}{\beta}\right)^{\kappa-1} e^{-(x / \beta)^\kappa},\quad x \geq 0, 
\]
where $\kappa$ is the shape parameter and $\beta$ is the scale parameter.
We change the scale parameter from $\beta_0 = 1$ to $\beta_1 = 0.6$, set the shape parameter $\kappa=1.5$, and the location shift $\mu_q = (\beta_0-\beta_1)\Gamma(1+1/\kappa)$.
\item {\it Gompertz distribution with scale and location shift:} The density of Gompertz distribution is
\[
f_{\text{Gompertz}}(x ; \kappa, \beta)=\frac{\kappa}{\beta} e^{\kappa+ x/\beta-\kappa e^{x/\beta}},\quad x \geq 0, 
\]
where $\kappa$ is the shape parameter and $\beta$ is the scale parameter.
We change the scale parameters from $\beta_0 = 1.5$ to $\beta_1 = 1$, set the shape parameter $\kappa=1$ and the location shift $\mu_q = (\beta_0-\beta_1)e^\kappa E_1(\kappa)$ with the exponential integral defined as
\[
E_1(z) = \int_z^\infty \frac{e^{-t}}{t} dt, \quad z>0.
\]
\end{enumerate}

\begin{figure}[t!]
\centering 
\begin{subfigure}[h]{0.45\linewidth}
\includegraphics[width=\linewidth]{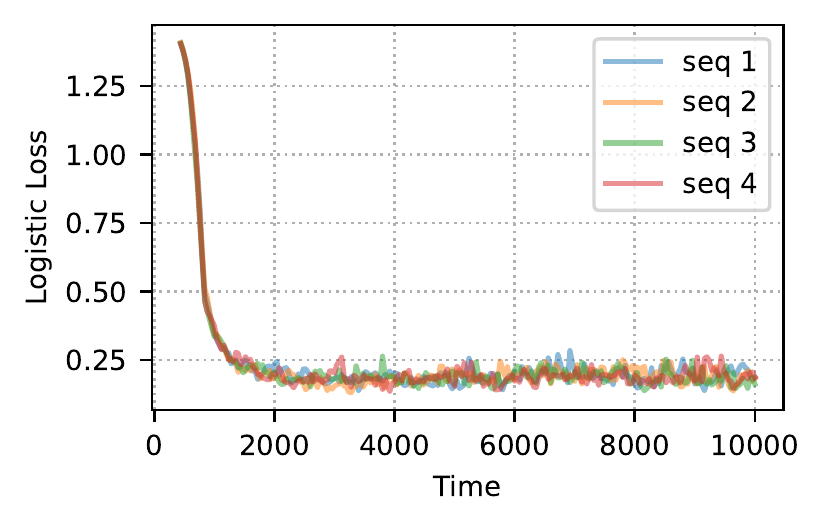}
\caption{Pre-change training logistic loss}
\end{subfigure}
\begin{subfigure}[h]{0.45\linewidth}
\includegraphics[width=\linewidth]{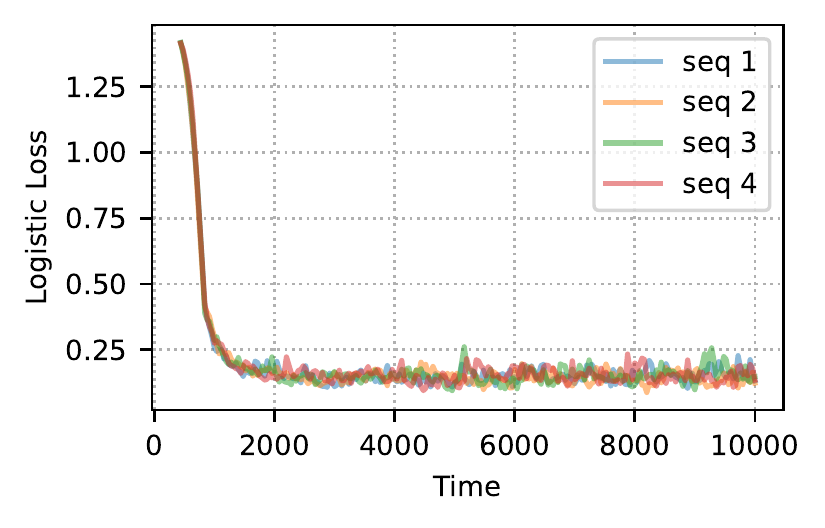}
\caption{Post-change training logistic loss}
\end{subfigure}
\caption{(a) Convergence of pre-change training logistic loss over time. (b) Convergence of post-change training logistic loss over time.}
\label{fig:Gaussian_mean_shift_training_convergence}
\end{figure}
\begin{table}[t!]
    \caption{Ten examples in Table~\ref{tab:simulation_summarize}, the estimated drifts $D$ and the increments $\eta_t-D$ are reported. The reported drift $D$ is the finite sample mean of pre-change $\eta_t$ over $500$ reference sequences: $\eta_t-D,\, t\le k$ and $\eta_t-D,\,t>k$ are finite sample means of pre- and post-change increments with removement of the estimated drifts $D$ over $400$ online sequences. The standard deviations are in parentheses.} 
    \label{tab:simulation-drifts-increments}
    \centering
    \begin{scriptsize}
\begin{tabular}{c ccccc}
\toprule
  
 $\times 10^{-4}$  & Gaussian (mean) & Gaussian (cov.) & Log Gaussian & GMM & 
     Chi-square \\
\midrule
$D$
 & -7.81 (155.042)
 & 0.23 (152.713)
 & 4.91 (127.739)
 & 3.61 (142.255)
 & 4.93 (130.959)
\\[1pt]
$\eta_t-D,\, t\le k$
 & 3.65 (146.189)
 & -1.45 (167.073)
 & -7.16 (131.945)
 & -10.02 (136.031)
 & 2.09 (141.910)
\\[1pt]
$\eta_t-D,\, t > k$
 & 40.59 (55.647)
 & 973.73 (256.498)
 & 589.27 (130.418)
 & 3313.88 (207.756)
 & 535.65 (132.599)
\\[1pt]
\bottomrule
\end{tabular}
\end{scriptsize}

\vspace{+10pt}

\begin{scriptsize}
\begin{tabular}{c c@{\hskip 15pt}c@{\hskip 15pt}c@{\hskip 15pt}c@{\hskip 15pt}c}
\toprule
  
 $\times 10^{-4}$  & Pareto & Exponential & Gamma & Weibull & 
     Gompertz \\
\midrule
$D$
& 10.62 (378.284)
& -12.02 (100.384)
& -3.28 (99.617)
& 1.20 (86.465)
& -5.25 (89.504)
\\[1pt]
$\eta_t-D,\, t\le k$
& -43.26 (497.288)
& -0.18 (113.002)
& 0.26 (87.107)
& -7.37 (94.489)
& -3.15 (91.676)
\\[1pt]
$\eta_t-D,\, t > k$
& 3712.11 (530.630)
& 882.27 (164.397)
& 308.23 (93.692)
& 1265.59 (339.828)
& 833.51 (246.964)
\\[1pt]
\bottomrule
\end{tabular}
\end{scriptsize}
\end{table}
\subsection{Additional results}
\noindent{\it Numerical justification.}
Figure~\ref{fig:Gaussian_mean_shift_training_convergence} demonstrates the training convergence of the separate training on pre- and post-change sequences of the neural networks. Multiple sequences are plotted simultaneously. The figure serves as a sanity check to validate the stable convergence of the neural network in the proposed scheme. 

\vspace{.1in}
\noindent{\it Numerical comparisons.} 
We tune the drifts $D$ in \eqref{eq:NNCUSUM_recursion} by training neural networks on 500 reference sequences sequentially and computing each sequence's temporal sample means of $\eta_t$. In this way, we can have a total of 500 empirical drifts. In the online change-point detection, we deploy the finite sample mean over 500 drifts in \eqref{eq:NNCUSUM_recursion}. On each online sequence, we compute the temporal pre- and post-change sample means of $\eta_t-D$ for $t\le k$ and $t>k$, respectively. Then, we yield 400 temporal means for pre- and post-change increments from 400 online sequences. 

Table~\ref{tab:simulation-drifts-increments} reports the finite sample means and standard deviations of (temporal means of) the drifts, pre- and post-change increments over the sequences. We observe that all drifts and pre-change increments are insignificant regarding a much larger standard deviation than its mean. Almost all post-change increments are significantly positive except for the Gaussian mean shift example, whose post-change increment is still much larger than the pre-change one. Table~\ref{tab:simulation-drifts-increments} is another justification that NN-CUSUM can work well over ten difficult examples due to the positive post-change $\eta_t$.

\end{document}